\documentclass[11pt]{article}
\usepackage[preprint]{acl}
\usepackage{times}
\usepackage{latexsym}
\usepackage[T1]{fontenc}
\usepackage[utf8]{inputenc}
\usepackage{microtype}
\usepackage{inconsolata}
\usepackage{graphicx}
\usepackage{subcaption}
\usepackage{booktabs}
\usepackage{multirow}
\usepackage{placeins}
\usepackage{float}
\usepackage{dblfloatfix}
\makeatletter
\@ifpackageloaded{lineno}{%
  \@ifpackagelater{lineno}{2026/02/10}{%
    \@ifpackagelater{lineno}{2026/03/07}{}{%
      \RemoveFromHook{build/column/before}[lineno]
      \AddToHook{build/column/after}[lineno]{%
        \setbox\@outputbox\vbox to\ht\@outputbox{%
          \boxmaxdepth\@maxdepth
          \protected@write\@auxout{}{%
            \string\def\string\@LN@column{\if@firstcolumn1\else2\fi}%
          }%
          \unvbox\@outputbox
        }%
      }%
    }%
  }{}%
}{}%
\makeatother
\usepackage{amsmath}
\usepackage{amssymb}
\usepackage{mathtools}
\usepackage{amsthm}
\usepackage[capitalize,noabbrev]{cleveref}
\crefname{appendix}{App.}{Apps.}
\Crefname{appendix}{App.}{Apps.}
\crefname{subappendix}{App.}{Apps.}
\Crefname{subappendix}{App.}{Apps.}
\crefname{subsubappendix}{App.}{Apps.}
\Crefname{subsubappendix}{App.}{Apps.}
\crefname{subsubsubappendix}{App.}{Apps.}
\Crefname{subsubsubappendix}{App.}{Apps.}
\usepackage{colortbl}

\newcommand{\stu}{\pi_{S}}
\newcommand{\tea}{\pi_{T}}

\title{Demystifying On-Policy Distillation:\\Where Does the Teacher's Advantage Live?}
\author{Yiping Li, Zhibo Hou, Fan Zhao, Zhiyu An, Wan Du \\
  University of California, Merced, USA \\
  \texttt{\{yipingli,zhou6,fanzhao,zan7,wdu3\}@ucmerced.edu}}
\begin{document}
\maketitle
\begin{abstract}
On-policy distillation (OPD) is increasingly studied as a cheaper alternative to reinforcement learning, yet it remains a form of knowledge distillation (KD) that relies on teacher supervision. We revisit OPD from the KD perspective by asking where the teacher's advantage lies along student-generated trajectories. Controlled token flips at teacher--student disagreements directly measure the accuracy gained from teacher corrections. Single-token flips yield a modest average gain on selected failed trajectories, whereas repeated flips raise accuracy from 28.4\% to 74.5\%, close to the teacher's 76.9\%. Gains continue to accumulate with more flips, but mass- and KL-based selection outperform random flips by 15--18 percentage points at counts of similar magnitude. Selective intervention remains useful after OPD, but greater argmax agreement does not consistently improve accuracy across model pairs. We advocate training schedules that learn more valuable teacher corrections while maintaining distributional stability.
Controlled flips and fork-resolution analysis evaluate OPD beyond accuracy by measuring which teacher preferences the student acquires and what further correction can recover.
\end{abstract}

\section{Introduction}
\label{sec:intro}

Reinforcement learning improves large language models after pretraining, but it is expensive \citep{deepseekr1,qwen3,dapo,respec}. On-policy distillation (OPD) offers a cheaper route by training the student on its own rollouts toward a teacher's next-token distribution \citep{gkd,opd_tm}. Reported results reach RL-level accuracy at about a tenth of the cost \citep{qwen3,opd_tm}. This efficiency has brought OPD into frontier post-training pipelines and attracted research interest \citep{deepseekv4,mimo,glm5,kimik3,nemotroncascade2,song2026survey}.

Yet OPD is commonly treated as an RL method because it serves as a cheaper substitute for RL in post-training. This view captures its role in the pipeline, but OPD still depends on a teacher and remains a form of knowledge distillation (KD). In conventional KD, soft labels preserve similarities that hard labels discard \citep{hinton2015}. For OPD, however, it remains unclear what in the teacher's next-token distributions contributes to better reasoning and where that contribution appears along the student's trajectory.

Prior work suggests that distillation quality depends on both what the teacher provides and what the student can learn \citep{dcd2022}. Therefore, OPD training outcomes alone do not distinguish the teacher's available advantage from the student's ability to acquire it. We first examine the teacher's contribution through interventions that hold the student's parameters fixed, then analyze how OPD checkpoints change the disagreement.

In this paper, we focus on \emph{forks}, positions at which the student and teacher diverge in their next-token preferences. We measure the teacher's advantage over the student through single and repeated token flips. A single flip replaces the student's token at one fork and lets the student continue. Dual-track decoding repeats this one-token intervention at selected forks as the trajectory develops.

We use these interventions to ask how much accuracy teacher corrections recover and whether selecting particular forks improves that recovery. Checkpoint analysis then asks which preferences OPD acquires and whether learning them produces better answers. Our contributions are the following.
\begin{itemize}\setlength{\itemsep}{0pt}
  \item \textbf{We develop an intervention-based evaluation of OPD.} Controlled flips measure the accuracy recoverable through teacher corrections, while checkpoint analysis measures which teacher preferences the student acquires. This separates the value of supervision from its learnability.
  \item \textbf{We characterize how teacher corrections improve student trajectories.} Gains accumulate through repeated flips, but mass- and KL-based selection recover more accuracy than random flips at intervention counts of similar magnitude. Teacher guidance is therefore broadly useful without making all disagreements equally valuable.
  \item \textbf{We identify a mismatch between token-level learning and trajectory-level improvement.} Across model pairs, fork resolution can increase while accuracy stagnates or declines, and selective intervention remains useful after training. Resolving disagreements alone cannot establish whether OPD turns teacher guidance into better answers.
\end{itemize}

\section{Related Work}
\label{sec:related}

\paragraph{On-policy distillation and its variants.}
OPD trains on student-generated trajectories with teacher supervision, often using reverse KL \citep{gkd,opd_tm}. TA-OPD retains high-teachability tokens, with a 10\% budget in its main experiments \citep{taopd}, while TIP emphasizes low-entropy tokens with high teacher--student divergence \citep{xu2026tip}. Prefix methods study shorter rollouts and supervision from early trajectory segments \citep{zhang2026prefix,zhou2026earlystop,liu2026suffixfade,zhang2026fullrollouts}.

Other methods modify supervision through advantage regulation or baselines \citep{wang2026demystifying,zhao2026poweropd,oh2026vopd}, support-aware matching, asymmetric objectives, or reward extrapolation \citep{fu2026revisiting,jia2026aopd,yang2026gopd}. EOPD adds forward KL at high-teacher-entropy positions \citep{jin2026entropy}, while TrOPD chooses the divergence using a trust region on the teacher-to-student ratio \citep{tropd}. TGPO supervises teacher-preferred tokens on student-generated prefixes without intervening in generation \citep{liu2026tgpo}.

Teacher intervention instead changes the rollout through reflection-triggered handoffs \citep{relayopd}, backtracking or prefix refinement \citep{wang2026motab,jiang2026trd}, or trust-region behavior blending during warm-up \citep{plyusov2026blending}. Other work changes the data or reference \citep{liu2026simpleopd,luo2026length,hou2026uniopd}, and \citet{song2026survey} survey OPD.

These rules encode different assumptions about where the teacher's advantage lies. They favor sparse positions, uncertain positions, prefixes, or reflection words, but they do not directly test whether the advantage is localized in any of them. We instead ask where the teacher's higher trajectory-level accuracy enters the student's trajectory.

\paragraph{Analyses of OPD.}
A smaller line of work studies OPD rather than extending it. \citet{rethinkopd} examine which teachers help which students by tracking how their distributions move during training. \citet{wang2026demystifying} study the role and failure modes of OPD through training curves across teachers and benchmarks. \citet{luo2026length} connect instability to the dynamics of the per-token advantage. These analyses take the training run as their object, so the outcome combines what the teacher offers with what the student learns. Our interventions isolate the teacher's contribution with fixed student parameters, and checkpoint comparisons separately track changes in token-level disagreement.

\paragraph{Assisted and guided decoding.}
Combining models during decoding has been used both as a method and as a lens for analysis. Speculative decoding lets a large model accept or replace tokens drafted by a smaller model, with speed as the objective \citep{leviathan2023,chen2023speculative}. Contrastive decoding and proxy tuning combine model predictions to improve the generated sequence \citep{li2023contrastive,liu2024proxy}. We use a lockstep teacher--student pairing as a measurement in which the student retains its sampled token unless the intervention selects the position. The number and location of teacher substitutions then become the quantities of interest.

\section{Preliminaries}
\label{sec:prelim}

\paragraph{On-policy distillation.}
Let $\stu$ be the student and $\tea$ the teacher. Given a prompt $P$, OPD trains on the student's own trajectories $x=(x_1,\dots,x_T)\sim\stu(\cdot\mid P)$ of length $T$. At every position $t$ the two models see the same prefix and propose a next token: $q_t=\stu(\cdot\mid P,x_{<t})$ is the student's distribution and $p_t=\tea(\cdot\mid P,x_{<t})$ the teacher's. The objective pays, at every position, a weighted price for being different from the teacher:
\par
\begingroup\small
\begin{equation}
\mathcal{L}=\mathbb{E}_{x\sim\stu}\!\sum_{t=1}^{T}w_t\mathrm{KL}_t,\quad
\mathrm{KL}_t=\mathrm{KL}(q_t\|p_t).
\label{eq:opd}
\end{equation}
\endgroup
where $w_t$ is the weight of position $t$, with $w_t\equiv1$ for uniform supervision. Summing over the whole vocabulary at every position is expensive, so our top-$k$ approximation keeps the tokens the student ranks highest, with $k=16$:
\par
\begingroup\small
\begin{equation}
\mathrm{KL}_t\approx\widehat{\mathrm{KL}}_t
=\!\!\sum_{x\in\mathrm{Top}_k(q_t)}\!\! q_t(x)\log\frac{q_t(x)}{p_t(x)}.
\label{eq:dhat}
\end{equation}
\endgroup
The gradient therefore reaches only tokens the student already considers plausible; a teacher token outside $\mathrm{Top}_k(q_t)$ is never pushed up directly.

\paragraph{Choice of divergence.}
\Cref{eq:opd} uses the reverse KL, which is mode-seeking: the student is charged for probability it places where the teacher places little. GKD also considers forward KL and Jensen--Shannon \citep{gkd}, while TrOPD uses a position-dependent mix of reverse and forward KL \citep{tropd}. Our experiments use reverse KL throughout.

\section{Setup}
\label{sec:setup}

\subsection{Models, Data, and Grading}
The student $\stu$ is Qwen3-1.7B in non-thinking mode; the teacher $\tea$ is Qwen3-4B-Instruct-2507 \citep{qwen3}. They share a tokenizer, so tokens compare position by position. Problems come from DAPO-Math-17k \citep{dapo}. A trajectory is correct when its final \texttt{Answer:}/\texttt{\textbackslash boxed} expression is equivalent to the reference. Unless stated otherwise both models sample at temperature $0.7$, top-$p$ $0.95$, and teacher-alone baselines use the same sampling. All runs use a fixed seed of 0; the hardware, software and decoding settings of every stage are listed in \cref{app:compute}.

\subsection{Fork Definitions and Scores}
\label{sec:concepts}

\paragraph{Fork.}
A \emph{fork} is a position where the student and teacher prefer different next tokens under the same prefix,
\begin{equation}
\arg\max q_t\neq\arg\max p_t,
\label{eq:fork}
\end{equation}
while the strength of their disagreement is measured separately below. A sampled student token can still equal the teacher's argmax at a fork, so a fork need not produce a token replacement.

\paragraph{Token mass.}
KL measures distribution-level disagreement but is unbounded and is not a metric \citep{coverthomas}. We also use \emph{token mass} to distinguish two directions of disagreement: the teacher assigning little probability to the student's token, and the student assigning little probability to the teacher's choice. For a token set $S$, these probabilities are
\begin{equation}
\mu^{T}_t(S)=\sum_{x\in S} p_t(x),\qquad \mu^{S}_t(S)=\sum_{x\in S} q_t(x),
\label{eq:AB}
\end{equation}
and $S$ can be a single token or a top-$k$ set, for example $\{x_t\}$ or $\mathrm{Top}_k(q_t)$. Similar quantities appear in prior work \citep{tropd,relayopd,taopd}.

\paragraph{Flip.}
A \emph{flip} replaces the student's token $x_t$ at a fork with a teacher token, either sampled from $p_t$ or chosen as its argmax. Each experiment specifies the replacement rule and continuation model. Changing one token does not make the subsequent continuation a teacher-generated trajectory.

\paragraph{Blind Forks.}
A fork is \emph{blind} when the teacher's preferred token lies outside the student's top-$k$ candidates, $\arg\max p_t\notin\mathrm{Top}_k(q_t)$, with $k=16$. The teacher's preferred token therefore has no explicit supervision term in the top-$k$ loss. A \emph{reverse-blind fork} instead has the student's token outside the teacher's top-$k$ candidates, $x_t\notin\mathrm{Top}_k(p_t)$.

\paragraph{Reflection words.}
We use the 13-word list of Relay-OPD \citep{relayopd}, including \emph{wait}, \emph{but}, and \emph{actually}, matched at the token level (\cref{app:reflect}). This operational definition includes words that can also occur outside reflection.

\begin{table*}[!t]
\centering\footnotesize\setlength{\tabcolsep}{3pt}
\caption{Single-flip gains over resampling and net-gain differences versus remaining positions, in pp. Each estimate is paired with its test at the same aggregation level. Share uses 5,901 eligible forks. Group tests are unadjusted. Full results and eligibility rules appear in \cref{tab:flipfullsignificance,app:stats_singleflip}.}
\label{tab:flipmain}
\begin{tabular*}{\textwidth}{@{\extracolsep{\fill}}lrrrrrrrrr@{}}
\toprule
 & & \multicolumn{4}{c}{Flip $-$ Resample} & \multicolumn{4}{c}{Net Gain $-$ Rest} \\
\cmidrule(lr){3-6}\cmidrule(lr){7-10}
 & & \multicolumn{2}{c}{Trajectory} & \multicolumn{2}{c}{Problem} & \multicolumn{2}{c}{Trajectory} & \multicolumn{2}{c}{Problem} \\
Fork Group & Share (\%) & Gain & $p_T$ & Gain & $p_P$ & $\Delta$ & $p_T$ & $\Delta$ & $p_P$ \\
\midrule
All Forks & 100.0 & +1.41\makebox[1.2em][l]{$^{***}$} & $<0.001$ & +1.22\makebox[1.2em][l]{$^{*}$} & 0.015 & -- & -- & -- & -- \\
\midrule
$\mu_t^S<0.05$ & 53.9 & +2.06\makebox[1.2em][l]{$^{***}$} & $<0.001$ & +1.83\makebox[1.2em][l]{$^{**}$} & 0.007 & +1.40\makebox[1.2em][l]{$^{***}$} & $<0.001$ & +1.35\makebox[1.2em][l]{$^{**}$} & 0.004 \\
Blind & 9.0 & +5.27\makebox[1.2em][l]{$^{***}$} & $<0.001$ & +4.27\makebox[1.2em][l]{$^{*}$} & 0.012 & +4.08\makebox[1.2em][l]{$^{***}$} & $<0.001$ & +3.31\makebox[1.2em][l]{$^{*}$} & 0.015 \\
Reflection/Contrast & 6.6 & +6.59\makebox[1.2em][l]{$^{*}$} & 0.045 & +5.11\makebox[1.2em][l]{} & 0.116 & +1.84\makebox[1.2em][l]{$^{**}$} & 0.008 & +3.84\makebox[1.2em][l]{} & 0.189 \\
\bottomrule
\end{tabular*}

\end{table*}

\subsection{Interventions}
\label{sec:interventions}

\paragraph{Single flip.}
One fork of an existing student trajectory is flipped and nothing else before it is changed. The teacher supplies a sampled segment of $L$ tokens, including the first replacement, and the student then writes the rest of the trajectory. At $L=\infty$ the teacher finishes the trajectory. The baseline independently resamples the student from the same prefix, including the fork token, to measure gain over another student continuation.

\paragraph{Dual-track decoding.}
We use dual-track decoding as a controlled intervention on student generation. Both models condition on the current prefix, with one forward pass each per step. At each fork selected by the flip condition, we replace only the student's next token with the teacher's argmax, $\arg\max p_t$, then return control to the student. At all other positions, the student samples from its own distribution. Replacements stop once an answer marker appears. We also test sampling the replacement from the teacher's distribution while keeping the same selection rule (\cref{tab:dualsampling}). This measures the combined effect of repeated token replacements along the resulting trajectory.

\section{Empirical Analysis}
\label{sec:claims}

We sample four trajectories per problem from the student before OPD training, using 1{,}000 DAPO-Math problems. The student and teacher each score the same 5.24M token positions to characterize their disagreement. The interventions then use subsets of these problems, specified below.\footnote{Throughout the paper, asterisks indicate statistical significance: $^{*}p<0.05$, $^{**}p<0.01$, and $^{***}p<0.001$. Test details and units of analysis are given in \cref{app:stats}.}

\subsection{Population Statistics}
\label{sec:c1}
The student and teacher prefer the same next token at most positions, with forks accounting for 13.93\% of the rollout. The top 20\% of positions ranked by reverse KL contribute 94\% of the loss (\cref{app:pool}). At about 63\% of profiled forks, the teacher's preferred token is the student's second or third choice (\cref{app:populationprofiles}). We test whether correcting disagreements improves final answers rather than assuming that loss concentration identifies useful corrections.

\subsection{Single Flip}
\label{sec:c2}
We first ask whether one teacher token can rescue a failed trajectory. We select 10 problems that the teacher solves 4/4 from the prompt and the student 0/4 to make the performance gap observable. We intervene at 5{,}926 forks across 40 failed trajectories and compare recovery with 16 independent student resamples per prefix. Single-token estimates retain changed-token samples at 5{,}901 forks; longer segments and continuation counts are specified in \cref{app:singleflip_settings}.

\subsubsection{How Much Does a Single Teacher Token Help?}
Single-token replacement improves success from 9.91\% to 11.31\%, a trajectory-weighted gain of 1.41 pp over resampling at the same position (95\% CI $[0.66,2.15]$, $p_T<0.001$). The problem-weighted gain is 1.22 pp ($p_P=0.015$, \cref{tab:flipmain}). Extending the teacher's segment increases the trajectory-weighted mean gain to 3.25 pp at $L=16$, 16.51 at $L=128$, and 79.68 when the teacher writes to the end (\cref{app:singleflipoverall}). Whether the inserted token is the teacher's argmax or a lower-ranked sample makes no detectable difference in the original A100 continuation runs ($p=0.27$). The gain also holds against a keep-token baseline that preserves the original student token and resamples only its continuation (\cref{app:keep}).

\subsubsection{Which Forks Benefit More from a Single Replacement?}
We also report reflection/contrast positions because they may mark a change in reasoning direction, defining them by the teacher's preferred next token. The average gain varies across fork groups (\cref{tab:flipmain}). Positions with $\mu_t^S<0.05$ and blind positions show larger gains than their complements, by 1.35 and 3.31 pp respectively under problem-level weighting, with both comparisons significant. The larger gain at blind forks mainly reflects lower student-resampling success rather than much higher replacement success (\cref{app:fliprates}). Reflection/contrast positions have a positive estimated gain over resampling, but neither this gain nor its difference from other forks is significant under the problem-level test. Full group comparisons across intervention lengths are reported in \cref{app:flipgroups}.

Despite these differences in average gain, the tested features predict only weakly whether an individual flip outperforms resampling. Combining all features yields a cross-validated AUC of 0.554 (\cref{app:prediction}). Group averages can favor a selection rule even when the features do not reliably distinguish which flips outperform resampling.

\subsubsection{Does Intervention Timing Matter?}
Later prefixes are harder to recover even when the teacher completes the continuation (\cref{app:singleposition}). Paired single-flip gains do not distinguish a position group from the remaining forks under both aggregation levels. Timing primarily describes recoverability rather than incremental token value.

\subsubsection{Can the Student Follow a Teacher Correction?}
A correction could lead into a continuation the student struggles to follow, but entropy approaches the unchanged-token control after about 40 tokens (\cref{app:afterflipdiagnostics}). On teacher trajectories, 73.4\% of fork runs contain only one fork and fewer than 1\% reach five tokens (\cref{app:teacherspacing}), suggesting local rather than sustained disagreement.

\subsection{Dual-Track Decoding}
\label{sec:c3}
We apply repeated flips on 200 randomly sampled problems with four trajectories each. Since flips can shorten or extend trajectories, we compare rules at flip counts of similar magnitude.

\begin{figure}[!htbp]
\centering
\includegraphics[width=\columnwidth]{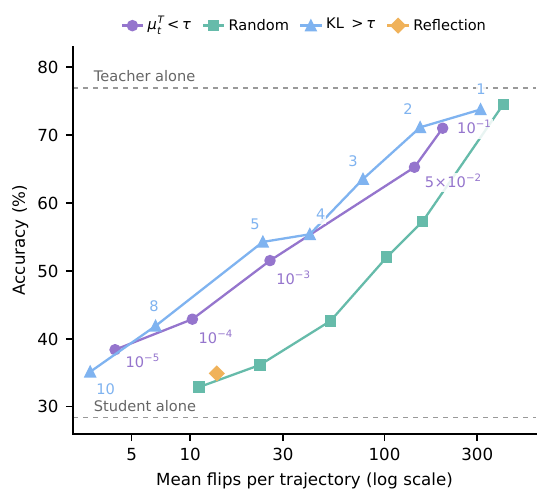}
\caption{Accuracy versus mean flips per trajectory. Point labels indicate selection thresholds. Full results are in \cref{tab:dualcomplete}.}
\label{fig:flipscaling}
\end{figure}

\subsubsection{How Much Do Repeated Flips Improve Accuracy?}
The teacher's contribution accumulates over repeated corrections rather than requiring the student to follow an entire teacher-generated trajectory (\cref{fig:flipscaling}). From a student baseline of 28.4\%, random flips raise accuracy to 36.1\% with about 23 flips per trajectory, 52.0\% with 103 flips, and 57.3\% with 157 flips. At about 408 flips, covering roughly 90\% of forks, accuracy reaches 74.5\%, compared with the teacher's 76.9\%. To check whether late teacher hints drive the gains, we stop intervention earlier and let the student finish alone. Gains remain roughly proportional to the preceding flips across exit positions (\cref{app:handoff}), supporting accumulated correction rather than late answer hints as an explanation for the improvement.

\subsubsection{Does Fork Selection Matter?}
Both $\mu_t^T$- and KL-based selection shift the accuracy curve upward relative to random flips (\cref{fig:flipscaling}). They achieve higher accuracy with flip counts of similar magnitude. For example, at about 23--26 flips per trajectory, the two rules outperform random selection by 15.4 and 18.1 pp, respectively (both $p<10^{-9}$). The KL curve is slightly higher than the $\mu_t^T$ curve over much of the tested range. Mass and KL are therefore useful for allocating repeated corrections despite their weak prediction of individual single-flip gains. Ablation experiments across model sizes, model families, and datasets show similar patterns (\crefrange{app:additionalpairs}{app:otherfamilies}).

\subsubsection{Do Reflection Tokens Provide a Particular Advantage?}
Reflection-based selection increases the student's subsequent reflection-word use and trajectory length, but does not significantly improve accuracy over random flips ($p=0.29$, \cref{tab:style}). In contrast, $\mu_t^T<10^{-5}$ achieves significantly higher accuracy than random flips with fewer interventions, without a comparable increase in reflection-word use. Its accuracy advantage over reflection-based selection is not significant ($p=0.07$). More visible reflection is therefore not sufficient evidence of more effective teacher guidance.

\begin{table}[!htbp]
\centering\footnotesize\setlength{\tabcolsep}{3pt}
\setlength{\tabcolsep}{1.8pt}
\caption{Reflection-word use, trajectory length, and accuracy under sparse intervention. Reflection counts exclude teacher-inserted tokens. Significance compares accuracy with random flips.}
\label{tab:style}
\begin{tabular}{lcccc}
\toprule
Flipped & Flips & \shortstack{Introduced\\Reflection Words} & Length & \shortstack{Accuracy\\(\%)} \\
\midrule
Student alone & 0 & 2.5 & 1,345 & 28.4 \\
\midrule
random 6\% & 11 & 3.2 & 1,506 & 32.9 \\
reflection words & \textbf{14} & \textbf{8.4} & \textbf{2,156} & 34.9 \\
$\mu^{T}_t<10^{-5}$\rlap{$^{\boldsymbol{**}}$} & 4 & 3.1 & 1,594 & \textbf{38.4} \\
\bottomrule
\end{tabular}

\end{table}

\subsubsection{Can Teacher Intervention Hurt the Student?}
Teacher intervention need not help on every problem, even when the teacher performs better overall. On problems the student solves but the teacher does not, intervention reduces student accuracy from 31.2\% to 18.8\% in a small subgroup of 16 trajectories (\cref{app:dualgroups}). This shows that teacher corrections can also move the student away from a correct answer. Sampled and argmax teacher replacements are compared in \cref{app:teachersampling}.

\subsection{After OPD}
\label{sec:training}
Having examined where the teacher's advantage lies, we next ask how much of it the student acquires through vanilla OPD. The main analysis uses Qwen3-4B-Instruct-2507 as teacher and Qwen3-1.7B as student on DAPO-Math. We extend checkpoint analysis to the 4B-to-0.6B and 8B-to-4B Qwen pairs, Llama 3.2 3B-to-1B, and Gemma 3 12B-to-4B. We also examine an instruction-following run of the 1.7B Qwen student evaluated on IFEval (\cref{app:additionalcheckpoint_settings}).

\begin{figure}[!htbp]
\centering
\includegraphics[width=\columnwidth]{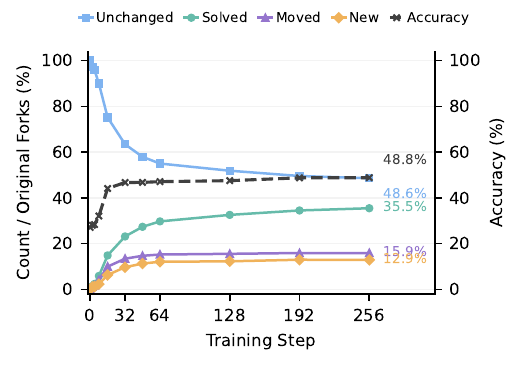}
\caption{Fork changes on pre-training prefixes (left, \% of original forks) and accuracy on each checkpoint's own rollouts (right).}
\label{fig:checkpointforktrends}
\end{figure}

\subsubsection{How Does Fork Resolution Change During Training?}
We track fork changes by scoring the same pre-training student trajectories with each checkpoint.
\begingroup
\setlength{\parskip}{0pt}
\begin{list}{\textbullet}{\setlength{\topsep}{4pt}\setlength{\partopsep}{0pt}\setlength{\itemsep}{0pt}\setlength{\parsep}{0pt}}
\item \textbf{Solved.} An original fork now agrees with the teacher.
\item \textbf{Unchanged.} An original fork retains the pre-training student's preference.
\item \textbf{Moved.} An original fork chooses neither the teacher's nor the pre-training student's token.
\item \textbf{New.} A position where the pre-training student and teacher agreed now disagrees with the teacher.
\end{list}
\endgroup

Fork-group proportions generally change smoothly during training, with few large reversals across model pairs. Solved forks increase and Unchanged forks decrease. In the main Qwen pair, Solved and New proportions settle near 35.5\% and 12.9\% of the original fork count by Final (\cref{fig:checkpointforktrends,tab:currentcheckpointforks}). The plateau may reflect limited capacity to acquire further teacher preferences.

However, accuracy does not consistently improve as more forks are resolved and can fluctuate while token agreement improves steadily. This suggests that OPD's effect on final answers depends on changes beyond agreement with the teacher at fork positions (\cref{app:checkpointpairs}).

\subsubsection{Which Forks Are More Learnable?}
\begin{figure*}[!t]
\centering
\includegraphics[width=\textwidth]{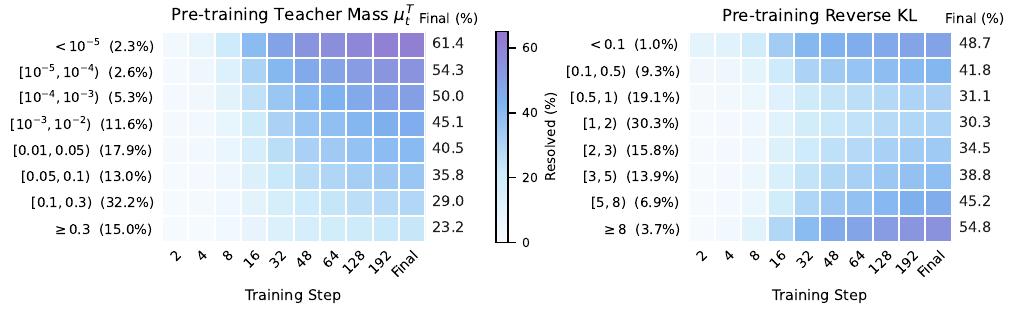}
\caption{Fork resolution by pre-training teacher mass and reverse KL. Colors show within-group rates, right-hand values show final rates, and parentheses show shares of original forks.}
\label{fig:forklearnabilitycurrent}
\end{figure*}

Fork resolution exhibits distinct mass- and KL-based patterns across model pairs rather than a universal ordering (\cref{fig:forklearnabilitycurrent,app:learnabilitypairs}). Runs using the same Qwen3-4B-Instruct-2507 teacher show similar patterns despite changes in student size or dataset (\cref{app:learnabilitypairs}). These patterns suggest that the teacher shapes which token choices are easier to learn, while the amount acquired varies with the student and training setting.

Blind forks are less often resolved despite their larger average single-flip gains. In the main Qwen pair, their resolution rate is 14.6\%, compared with 37.0\% for non-blind forks, consistent with the top-16 loss lacking an explicit supervision term for the teacher's preferred token.

\subsubsection{How Much Does Teacher Intervention Help after OPD?}
\begin{figure*}[!t]
\centering
\includegraphics[width=\textwidth]{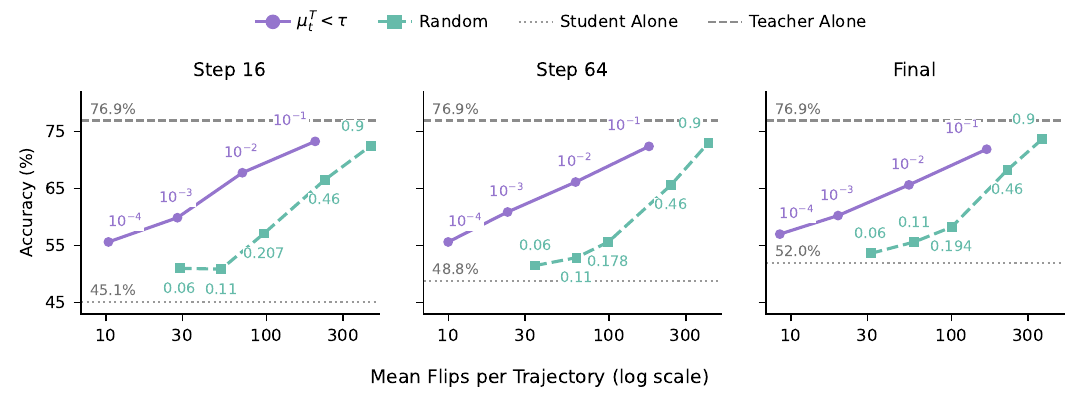}
\caption{Mass selection and random flips at three OPD checkpoints. Labels give mass thresholds or random probabilities. Gray lines mark standalone accuracy. Each point uses 800 trajectories.}
\label{fig:afteropdmassrandom}
\end{figure*}

\label{sec:afteropdintervention}
Mass-based selection remains more effective than random flips at low intervention counts after OPD (\cref{fig:afteropdmassrandom}). Its curves rise steeply and then flatten, while random flips require more interventions to reach similar accuracy. Applying the same mass threshold brings checkpoints with different starting accuracies to similar performance, suggesting that training has not exhausted the useful teacher corrections identified by this rule.

\subsubsection{Which Remaining Forks Are Useful Intervention Targets?}
\begin{figure*}[!t]
\centering
\includegraphics[width=\textwidth]{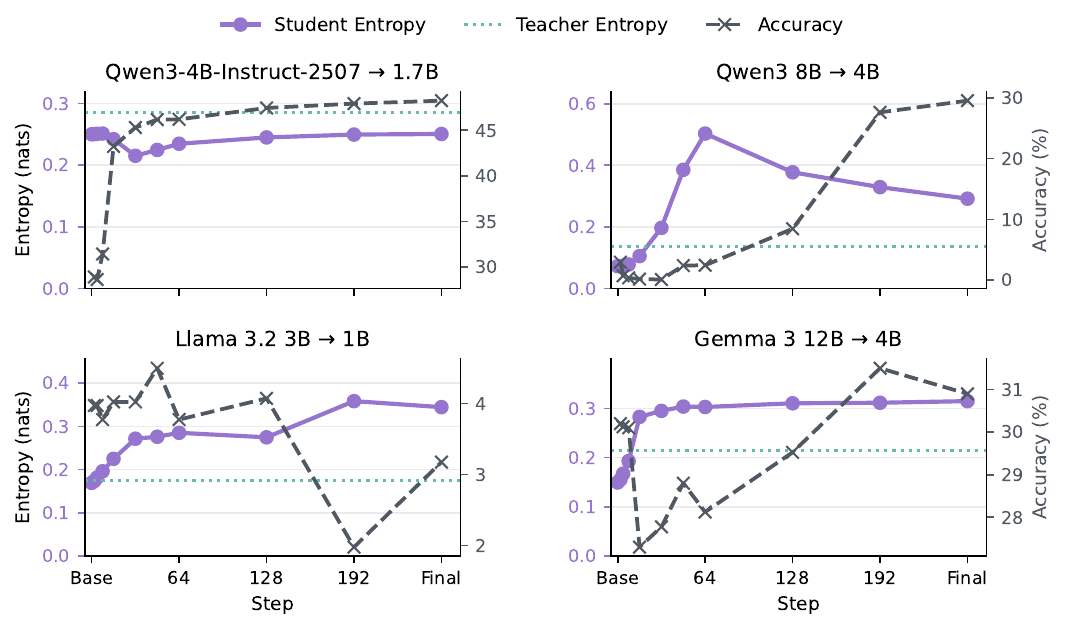}
\caption{Entropy on fixed pre-training prefixes (purple, left axes) and accuracy on checkpoint rollouts (gray, right axes). Teacher entropy is dotted. Post-answer repetition is excluded. All six settings appear in \cref{app:afteropdentropy}.}
\label{fig:afteropdentropymain}
\end{figure*}

\label{sec:afteropdforktargets}
We ask whether intervention is more useful at disagreements that persist through training or those introduced by training. At each fork encountered during generation, we also check the pre-training student's preferred token. If it agrees with the teacher, the fork is New. Otherwise, the fork is Unchanged if the trained student retains its original preference, or Moved if it chooses a different token. We compare these group-based interventions with random flips at counts of similar magnitude.

Prioritizing New or Moved forks does not produce a consistent accuracy advantage across model pairs (\cref{app:afteropdforkgroups}). In the main Qwen pair, New selection underperforms random controls, while Moved offers no detectable advantage. In Gemma and Llama, neither group significantly improves accuracy over unassisted decoding. Unchanged forks still provide useful corrections, but do not outperform random flips in the Qwen comparisons (\cref{tab:afteropdforkgroups}). Newly introduced or changed preferences are therefore weak signals of where further teacher correction will help.

\begin{table}[!htb]
\centering\footnotesize\setlength{\tabcolsep}{3pt}
\caption{Fork-group interventions at step 16. F/T is flips per trajectory and F/F is the percentage of forks flipped. $p$ compares accuracy with Random in each block (\cref{app:stats_dualtrack}).}
\label{tab:afteropdforkgroups}
\begin{tabular*}{\linewidth}{@{\extracolsep{\fill}}lrrrr@{}}
\toprule
Fork Group & F/T & F/F & Acc. (\%) & $p$ \\
\midrule
Random & 96.6 & 20.4 & 57.13 & -- \\
Old & 100.0 & 19.7 & 56.75 & 0.837 \\
Unchanged & 94.5 & 18.5 & \textbf{57.63} & 0.761 \\
\midrule
Random & 28.9 & 5.9 & \textbf{51.00} & -- \\
New & 31.4 & 6.8 & 44.88\rlap{$^{***}$} & $<0.001$ \\
\midrule
Random & 52.2 & 10.8 & 50.88 & -- \\
Moved & 45.7 & 9.0 & \textbf{51.13} & 0.885 \\
\bottomrule
\end{tabular*}

\end{table}

\subsubsection{Why Might Fork Resolution Fail to Improve Accuracy?}
As shown earlier, repeated flips rescue trajectories and generally recover more accuracy as more corrections are applied. Yet OPD can resolve more forks without improving accuracy, and some runs deteriorate. Fork resolution tracks changes in the preferred token, while generation depends on the full distribution. Entropy remains stable in the main Qwen pair, while late gains in the 8B-to-4B pair accompany declining entropy (\cref{fig:afteropdentropymain}).

Even where argmax agreement is preserved, entropy rises and the shared token's probability falls in Llama, Gemma, and IFEval (\cref{tab:nonforkdistribution}). The student therefore becomes less likely to sample that token despite retaining it as its preferred choice. We suspect that training acquires useful corrections while destabilizing choices elsewhere. A good OPD training schedule should help the student resolve more valuable forks while maintaining entropy stability (\cref{app:afteropdentropy}).

\section{Conclusion}\label{sec:discussion}

We systematically study OPD from the KD perspective by separating the value of teacher corrections from the student's ability to learn them. Evaluating OPD requires measuring both the teacher preferences acquired during training and the accuracy still recoverable through controlled corrections.

Repeated flips recover much more accuracy than a single intervention, and gains continue to accumulate as more corrections are applied. Mass- and KL-based selection allocate repeated corrections more effectively than random flips at counts of similar magnitude, despite weak prediction of individual single-flip gains.

Across model pairs, fork resolution follows smoother trends than accuracy and often approaches a plateau. In the main pair, only 35.5\% of original forks are resolved by the final checkpoint. Learnability patterns differ across teachers, and blind forks in the main pair are less often resolved despite their larger intervention gains. Intervention value therefore cannot substitute for evidence of learnability. Selective intervention remains useful after OPD, showing that saturated training accuracy need not mean that the teacher's useful guidance has been exhausted.

Greater token agreement nevertheless does not consistently produce better answers. Accuracy declines commonly accompany rising entropy across adjacent checkpoints. In several runs, entropy also rises where argmax agreement is preserved and the shared preferred token's probability falls. These changes motivate our hypothesis that OPD acquires useful corrections while destabilizing choices elsewhere. One direction is selective OPD that downweights New and Moved forks and places more supervision on Unchanged forks. A curriculum could further organize these corrections by their value and learnability, adjusting the pace of supervision to acquire more useful teacher preferences while maintaining distributional stability.

\clearpage
\section*{Limitations}

Our experiments cover several teacher--student pairs from the Qwen, Llama, and Gemma families, but the tasks are limited to mathematics and instruction following. Token-level interventions require a shared tokenizer within each pair. The single-flip study deliberately selects problems that the teacher solves and the student fails, making teacher corrections easier to observe. Its average gains therefore characterize this selected setting rather than randomly sampled problems.

We separate OPD into two questions: which teacher corrections improve student trajectories, and how the student learns those corrections. Controlled interventions measure the value of corrections, while checkpoint analysis characterizes which teacher preferences are acquired on fixed pre-training prefixes. Determining what makes valuable corrections learnable remains a question for further work. Our proposed training directions still require evaluation of whether they improve learning while preserving distributional stability.

\bibliography{refs}
\clearpage
\appendix
\crefalias{section}{appendix}
\crefalias{subsection}{subappendix}
\crefalias{subsubsection}{subsubappendix}
\newcommand{\appendixsection}[1]{\section{#1}}
\makeatletter
\setlength{\@fptop}{0pt}
\setlength{\@fpsep}{12pt}
\setlength{\@fpbot}{0pt plus 1fil}
\setlength{\@dblfptop}{0pt}
\setlength{\@dblfpsep}{12pt}
\setlength{\@dblfpbot}{0pt plus 1fil}
\makeatother
\setcounter{topnumber}{3}
\setcounter{dbltopnumber}{3}
\renewcommand{\topfraction}{0.95}
\renewcommand{\dbltopfraction}{0.95}
\renewcommand{\textfraction}{0.05}
\renewcommand{\floatpagefraction}{0.75}
\renewcommand{\dblfloatpagefraction}{0.75}
\appendixsection{Experimental Settings}\label{app:compute}
\subsection{Implementation}
Intervention and scoring runs use NVIDIA A100 and L40S GPUs on a Slurm cluster, with one process per GPU and both models in bfloat16. Sampling uses vLLM 0.11.0. Dual-track decoding, teacher forcing, and scoring use HuggingFace transformers 4.56.1 and PyTorch 2.8.0 with SDPA attention. Score log-softmax operations use float32. Figures and tables are generated from saved arrays or audited summaries.

All vLLM engines and samplers use seed 0. Dual-track decoding uses a PyTorch generator seeded from the seed, batch index, and shard. Random-flip controls use a NumPy generator with seed 0.

\subsection{Population Statistics}\label{app:population_settings}
For \cref{sec:c1}, we use 1{,}000 problems with 4 student rollouts each, temperature $1.0$, top-$p$ $1.0$, and at most 3{,}072 new tokens, following the rollout settings of \citet{rethinkopd}. The two next-token distributions at every position are scored with a top-16 truncation.

\subsection{Single Flip}\label{app:singleflip_settings}
Each fork receives four sampled teacher segments of length $L\in\{1,16,128,\infty\}$ and four student continuations per finite segment. Success is the fraction of continuations ending correct. The 16 independent student resamples per fork yield 94{,}816 baseline continuations. At $L=1$, we retain only teacher paths changing the original student token. All four teacher samples repeat that token at 25 positions, leaving 5{,}901 eligible forks.

At each of the 5{,}926 forks in \cref{sec:c2}, the teacher samples 4 continuations, temperature $0.7$, top-$p$ $0.95$, at most 8{,}192 tokens in an 8{,}192-token context. The $L\in\{1,16,128\}$ segments are prefixes of these, and the student writes 4 continuations of at most 3{,}072 tokens after each. For the reported $L=1$ results, we rerun student continuations on NVIDIA L40S GPUs, reusing the same four teacher replacement tokens per fork and drawing four continuations per token. The resample arm also runs on L40S, drawing 16 student continuations directly from each fork prefix. Both arms use the same vLLM pipeline, temperature $0.7$, top-$p$ $0.95$, a 3{,}072-token generation cap, and a 12{,}288-token context limit, and are paired by position. Longer-segment outcomes and token-level diagnostics retain the original A100 runs.

\subsection{Dual-Track Decoding}\label{app:dualtrack_settings}
For \cref{sec:c3}, both models decode in lockstep in transformers, temperature $0.7$, top-$p$ $0.95$, at most 8{,}192 tokens, with the answer-zone freeze of \cref{sec:interventions}. The student-alone and teacher-alone rows of \cref{tab:dual} are 4 vLLM samples per problem with the same settings and the same cap. The teacher pass@4 behind \cref{tab:cells} uses the 8{,}192-token cap; an earlier 4{,}096-token run truncated 38\% of the teacher's answers and was discarded.

Synchronous decoding uses KV caches and batch size 6. Before each replacement decision, we check the last 12 generated tokens for the answer-zone freeze. Each trajectory records token authorship, the student's sampled token, both argmaxes, and $A$ and $B$.

The additional model-pair and dataset experiments use chunked vLLM decoding. The student generates a chunk of tokens, which the teacher scores under the same prefixes. We accept tokens before the first selected fork, replace its token, discard the remaining chunk, and resume from the corrected prefix. Chunk size controls how many student tokens are generated and scored per round, not the length of teacher intervention. These experiments use chunks of 32 tokens, except for the 4B-to-0.6B KL runs, which use 8.

\subsubsection{Reflection-Word List}\label{app:reflect}
We use Relay-OPD's published list of \emph{wait, but, hmm, actually, hold, however, yet, oh, alternatively, no, ah, oops, well}. Matching ignores case, leading whitespace, and attached punctuation, yielding 79 Qwen3 token ids. Split forms such as \texttt{Wait} followed by punctuation still match the word token. We retain the full list, including words that also have non-reflective uses in mathematics, such as ``no solution'' and ``well-defined''.

\subsubsection{Additional Model Pairs}\label{app:additionalpair_settings}
We also use Qwen3-4B-Instruct-2507 to Qwen3-0.6B and Qwen3-8B to Qwen3-4B-Base on the same 200 problems, with four trajectories per problem, temperature 0.7, top-$p$ 0.95, an 8{,}192-token generation cap, and a 12{,}288-token context limit. Teacher replacements use argmax and stop at the answer-zone boundary. Reverse KL uses the student's top-16 tokens plus a residual-probability bucket. Each pair has five mass thresholds, five KL thresholds, and seven random probabilities, all with 800 completed trajectories.

\subsubsection{IFEval}\label{app:ifeval_settings}
We use Qwen3-4B-Instruct-2507 as teacher and Qwen3-1.7B as student on all 541 official IFEval prompts, with four trajectories per prompt. Standalone baselines and interventions use temperature 0.7, top-$p$ 0.95, a 6{,}144-token generation cap, and an 8{,}192-token context limit, with thinking disabled. We report official strict prompt accuracy, requiring all instructions in a prompt to be satisfied. Interventions use teacher argmax replacements and no answer-zone freeze. Each of the five mass thresholds, five KL thresholds, and seven random probabilities has 2{,}164 completed trajectories. Fork coverage divides total flips by total argmax disagreements on each policy's generated trajectories.

\subsubsection{Llama Model Pair}\label{app:llama_settings}
We use Llama-3.2-3B-Instruct as teacher and Llama-3.2-1B-Instruct as student on the same 200 DAPO-Math problems. Each of five mass thresholds, five KL thresholds, and seven random probabilities has 800 completed trajectories. Sampling uses temperature 0.7 and top-$p$ 0.95, with an 8{,}192-token generation cap and a 12{,}288-token context limit. Interventions use teacher argmax replacements and the answer-zone freeze. Reverse KL uses the student's top-16 tokens plus a residual-probability bucket.

\subsubsection{Gemma Model Pair}\label{app:gemma_settings}
We also evaluate Gemma-3-12B-it as teacher and Gemma-3-4B-it as student on the same 200 DAPO-Math problems, with four trajectories per problem. Sampling uses temperature 0.7 and top-$p$ 0.95, with an 8{,}192-token generation cap and a 12{,}288-token context limit. We compare mass-based, KL-based, and random fork selection using teacher argmax replacements and the answer-zone freeze. Reverse KL uses the student's top-16 tokens plus a residual-probability bucket.

\subsection{Handoff}\label{app:handoff_settings}
For each of the 800 dual-track trajectories, we retain the prefix up to a fraction $f\in\{0.25,0.5,0.75,0.9\}$ of its length before the answer zone and sample two student continuations. We also include $f=1$, where the student writes only the answer line, and the student-alone baseline at $f=0$.

For \cref{app:handoff}, the student continues each cut trajectory in vLLM, with 2 continuations per cut, temperature $0.7$, top-$p$ $0.95$, and at most 8{,}192 new tokens within a 12{,}288-token context.

\subsection{After OPD}\label{app:afteropd_settings}
We train Qwen3-1.7B with vanilla OPD using Qwen3-4B-Instruct-2507 as teacher. The DAPO-Math training data exclude the 1{,}000 evaluation problems. The current analysis uses checkpoints at steps 2, 4, 8, 16, 32, 48, 64, 128, 192, and the final step 256.

\subsubsection{Fork Resolution and Accuracy}\label{app:forkresolution_settings}
Each checkpoint scores all 5{,}237{,}483 positions from the same 4{,}000 pre-training student trajectories. Position identifiers align checkpoint scores with the saved pre-training student and teacher preferences. All 729{,}574 original argmax forks are included, even if the sampled student token equals the teacher's argmax. Solved, Unchanged, and Moved partition this set using the definitions in \cref{sec:training}. New forks lie outside the original set.

Accuracy uses each checkpoint's own 4{,}000 rollouts on the evaluation problems, with four samples per problem, temperature 1.0, top-$p$ 1.0, and an 8{,}192-token generation cap. The pre-training accuracy uses the same settings. Unlike accuracy, fork changes are measured on fixed pre-training prefixes.

\subsubsection{Fork Resolution by Pre-Training Scores}\label{app:learnability_settings}
We group the original forks using their saved pre-training scores. Teacher mass $\mu_t^T$ is the teacher's probability of the base student's argmax, and reverse KL uses the full vocabulary on the same prefix. Group membership is fixed across checkpoints. Resolution rates divide the number resolved at each checkpoint by the original count within that group. These are pooled position-level rates, not problem-weighted estimates.

\subsubsection{Teacher Intervention after OPD}\label{app:afteropdintervention_settings}
We repeat dual-track decoding at steps 16, 64, and the final checkpoint on the same 200-problem subset, using four trajectories per problem at temperature 0.7 and top-$p$ 0.95, with an 8{,}192-token cap. The mass thresholds are $10^{-4}$, $10^{-3}$, $10^{-2}$, and $0.1$. Random-flip probabilities are $0.06$, $0.11$, $0.46$, and $0.90$, supplemented by the approximately 100-flip random controls described below. Standalone checkpoint baselines use the same sampling settings. \Cref{fig:afteropdmassrandom} includes only complete 800-trajectory runs.

The additional Final-checkpoint KL interventions in \cref{app:afteropdfinalkl} use the student's top-16 tokens plus a residual-probability bucket. The 0.6B student uses the same 200 mathematics problems and decoding limits. IFEval uses all 541 official prompts, with a 6{,}144-token generation cap and an 8{,}192-token context limit. Both settings use temperature 0.7 and top-$p$ 0.95, with four trajectories per problem or prompt.

\subsubsection{Online Fork-Group Selection}\label{app:afteropdgroups_settings}
The pre-training student, trained student, and teacher score each current prefix to determine its fork group online. This differs from the fixed-prefix composition analysis. These interventions use the same checkpoints and sampling settings as \cref{app:afteropdintervention_settings}.

Old, Unchanged, and Random policies use probabilities calibrated toward 100 flips per trajectory. New and Moved policies intervene at every eligible fork, yielding fewer flips. At step 16, New is compared with random probability 0.06 and Moved with 0.11. At steps 64 and 256, both are compared with the approximately 100-flip random control. Step-64 New and Moved runs combine the initial shard with three further shards using the same probability of one, for 800 distinct problem--sample pairs per policy. All realized counts are reported in \cref{tab:afteropdforkgroupsfull}.

\subsubsection{Additional Checkpoint Evaluations}\label{app:additionalcheckpoint_settings}
We evaluate the additional Qwen, Llama, and Gemma checkpoints on 1{,}000 DAPO-Math problems with four rollouts per problem. The IFEval checkpoints use the 541 official prompts with four rollouts per prompt. These evaluations use temperature 0.7 and top-$p$ 0.95, with response caps of 8{,}192 tokens on DAPO-Math and 6{,}144 on IFEval. The supplementary 1.7B checkpoint curve also uses these sampling settings, separately from the temperature-1.0 evaluation in \cref{tab:currentcheckpointforks}. Fork changes use each pair's own fixed pre-training trajectories and original fork count. We do not pool positions across pairs. Mass and reverse-KL groups retain their pre-training membership at every checkpoint. Fixed-prefix fork statistics are complete at all ten checkpoints for every pair.

The additional Gemma and Llama fork-group interventions use 200 DAPO-Math problems with one rollout per problem at steps 16, 64, and 256. Each policy intervenes at every eligible fork in its selected group, at temperature 0.7 and top-$p$ 0.95. The response cap is 8{,}192 tokens and the context cap is 12{,}288. Intervention stops at the configured answer marker. Resolved positions already share the teacher's argmax and are not eligible for an argmax flip. These runs report the realized counts without budget calibration. Student-alone controls average four rollouts per problem on the same 200 problems at the same temperature and top-$p$. Llama uses the corresponding no-intervention runs from the same decoder. Gemma uses the 200-problem subset of its checkpoint rollouts, with a 9{,}216-token context cap. We verify that every selected prompt permits the full 8{,}192-token response cap.

The IFEval fork-group interventions use the same 541 prompts and four trajectories per prompt at steps 16, 64, and 256. Their no-intervention controls use the same sampling settings. Group policies are calibrated using recorded teacher insertions. We verify that each recorded insertion changes the student's preferred token in these completed runs, so the reported counts are actual flips.

For the entropy analysis in \cref{app:afteropdentropy}, we score each pair's fixed pre-training trajectories with the base student and every trained checkpoint. We compute full-vocabulary entropy from untempered model probabilities and average over token positions. DAPO-Math includes 4{,}000 trajectories per pair and IFEval includes 2{,}164. Scoring stops at the first complete answer when an answer boundary is found, excluding subsequent repetition. Accuracy uses each checkpoint's own rollouts at temperature 0.7 and top-$p$ 0.95.

\appendixsection{Statistical Tests}\label{app:stats}
The tests distinguish whether teacher intervention helps from whether one group or selection rule helps more than its comparator. Positions within a trajectory and trajectories from the same problem can be correlated, so we aggregate differences before testing rather than treating every fork as an independent observation. All reported $t$-tests are two-sided. Group-comparison $p$-values are unadjusted for multiple testing and interpreted as exploratory. Asterisks refer to the comparison specified in each table.

\subsection{Single-Flip Tests}\label{app:stats_singleflip}
We ask whether a teacher replacement improves recovery beyond another student attempt from the same prefix. At each fork, we subtract the success rate of 16 independent student resamples from the teacher-intervention success rate. Comparing continuations from the same prefix controls for how recoverable that position already is without the teacher. At $L=1$, only teacher paths changing the original token contribute, leaving 5{,}901 of 5{,}926 positions eligible. All teacher paths contribute at longer lengths.

\paragraph{Trajectory-Level Gain ($p_T$).}
To estimate the average gain across failed trajectories, we average paired position gains within each trajectory and apply a one-sample $t$-test against zero to the 40 trajectory means ($p_T$). Each trajectory receives equal weight. Reported single-flip confidence intervals use the Student-$t$ distribution.

\paragraph{Problem-Level Gain ($p_P$).}
To assess the gain across problems while aggregating their correlated trajectories, we pool eligible position gains within each problem and test the 10 problem means against zero ($p_P$). Each problem receives equal weight, so this test changes both the unit of analysis and the weighting relative to $p_T$.

\paragraph{Bootstrap Check.}
We use 2{,}000 trajectory-bootstrap replicates to check the sensitivity of the overall gain interval to the Student-$t$ construction. Each replicate resamples whole trajectory means rather than individual forks.

\paragraph{Fork-Group Differences.}
A positive gain over resampling does not establish that a group is more useful than other forks. We therefore compare the group's mean paired gain with the mean paired gain at the remaining positions. Within each trajectory or problem, we subtract the latter from the former and test these contrasts against zero. Each included unit must contain at least three eligible positions on each side. A positive significant contrast indicates a larger average teacher-intervention gain in that group, rather than merely a higher absolute recovery rate. Full estimates and tests, including nonsignificant comparisons, appear in \cref{app:flipgroups,app:fliprates}. Predicting the benefit of an individual flip is a separate question, addressed in \cref{app:prediction}.

\subsection{Dual-Track Tests}\label{app:stats_dualtrack}
Each policy generates four trajectories on the same 200 randomly sampled problems. For each problem, we subtract the comparator's mean correctness from the tested policy's mean correctness, then apply a one-sample $t$-test against zero to the 200 differences. This is equivalent to a paired $t$-test on the two sets of problem means. Pairing is over problems, not individual trajectories, because the policies generate different continuations.

\paragraph{Fork Selection versus Random Flips.}
This comparison asks whether choosing forks by a feature improves accuracy over random intervention. We apply the problem-paired test above to the selected policy and its random control. Controls approximate intervention coverage or counts, as specified for each experiment, rather than matching the exact realized number of flips. Random-control settings and comparison results for the original model pair are reported in \cref{app:dualmainresults}.

\paragraph{Teacher Sampling versus Argmax.}
This comparison asks whether sampling a teacher token changes accuracy relative to taking the teacher's argmax under the same selection rule. We apply the same problem-paired test to the two complete policies (\cref{tab:dualsampling}).

\paragraph{Fork-Group Selection after OPD.}
This comparison asks whether a fork's training history identifies useful intervention targets. We compare each group-selection policy with its designated random control using the same problem-paired test. Control choices and realized counts are specified in \cref{app:afteropdgroups_settings,app:afteropdforkgroups}.

\paragraph{Wilcoxon Check.}
For the original random-control comparisons, we also apply a Wilcoxon signed-rank test to the problem-level differences. This checks sensitivity to the paired $t$-test's treatment of those differences and gives the same significance conclusions.

\paragraph{Welch Check.}
For the same comparisons, a Welch $t$-test compares the 800 trajectory outcomes from each policy without pairing individual trajectories. This alternative aggregation also gives the same significance conclusions, while the problem-paired test remains the reported comparison.

\subsection{Handoff Tests}\label{app:stats_handoff}
\paragraph{Constant-Gain Residuals.}
This test asks whether accuracy at each teacher exit point departs from a constant-gain-per-flip model. Predicted accuracy at exit fraction $f$ interpolates between the problem's student-alone rate and the trajectory's accuracy at $f=1$, weighted by the fraction of flips made before $f$. We subtract this prediction from observed recovery and test the mean residual against zero over the 800 trajectories, then repeat the test after averaging within each problem.

\paragraph{Exit-Position Trend.}
This test asks whether departures from the constant-gain model vary systematically with the teacher's exit position. We fit a residual slope against $f$ within each trajectory and test the mean slope against zero. Results for both handoff tests appear in \cref{app:handoff}.

\appendixsection{Population Statistics}\label{app:pool}
\label{app:populationprofiles}
\begin{figure}[!htbp]
\centering
\includegraphics[width=\linewidth]{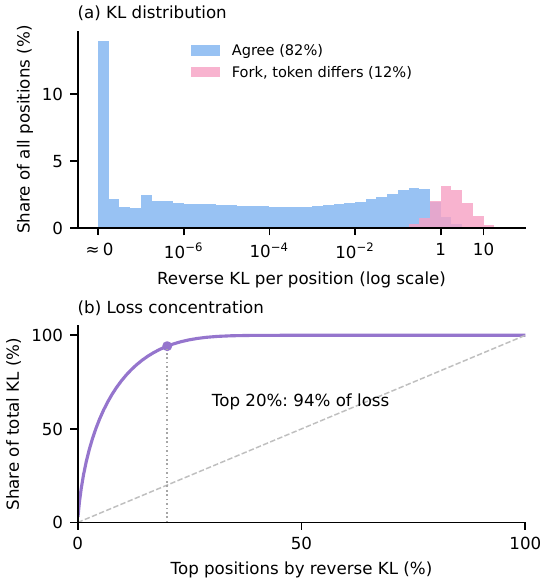}
\caption{Reverse KL on 5.24M base-model positions. (a) Distributions where both argmaxes and the sampled token agree, and at forks where the sampled token differs from the teacher's argmax. Bars use all positions as the denominator. Other sampling outcomes are omitted. (b) Cumulative loss over all positions ranked by reverse KL.}
\label{fig:population}
\end{figure}

\begin{figure*}[!tbp]
\centering
\includegraphics[width=0.90\linewidth]{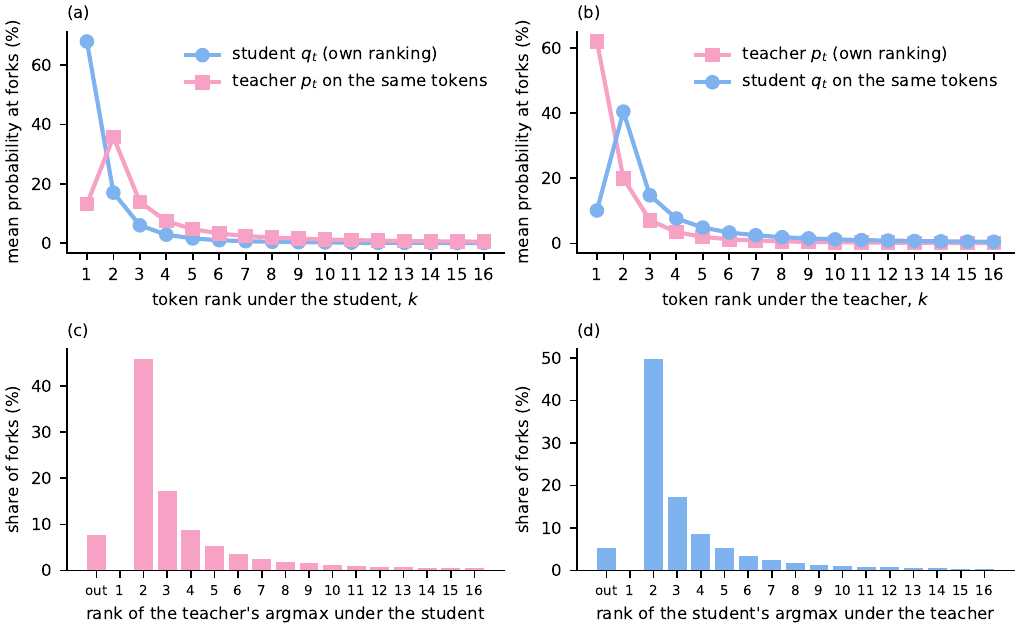}
\caption{Token choices at 640k argmax forks where the sampled token also differs from the teacher's choice. (a,b) Mean probabilities under student and teacher token rankings. (c,d) Rank of the other model's argmax, with ``out'' denoting a token outside the top-16.}
\label{fig:forkprofile}
\end{figure*}
The profiles use forks where the sampled student token also differs from the teacher's argmax, covering 12.23\% of all positions. A sampled-token mismatch alone need not be a fork, since sampling can select a different token even when both argmaxes agree. Loss concentration in \cref{fig:population} uses all positions without this filtering. Most probability mass lies on the top two or three candidates of either model. The teacher's argmax is the student's second choice at 46\% of profiled forks and its third at 17\%. In either ranking direction, the other model's argmax falls outside the top-16 at 7--8\% of these forks. These profiles distinguish disagreements over highly ranked alternatives from blind forks where the teacher's preferred token is outside the student's top-16.

\appendixsection{Single-Flip Gains}\label{app:fliptable}
\subsection{Overall Gains by Intervention Length}\label{app:singleflipoverall}\label{app:singlefliprank}
\begin{table}[!htbp]
\centering\footnotesize\setlength{\tabcolsep}{2pt}
\caption{Overall paired gains over the complete resample baseline. Effects and Student-$t$ 95\% intervals are trajectory-weighted, in pp. $p_T$ tests trajectory means and $p_P$ tests problem means. $n$ is the number of eligible positions (\cref{app:stats_singleflip}).}
\label{tab:flipoverall}
\begin{tabular*}{\linewidth}{@{\extracolsep{\fill}}lrrrrr@{}}
\toprule
$L$ & $n$ & Gain (pp) & 95\% CI & $p_T$ & $p_P$ \\
\midrule
1 & 5901 & +1.41 & [0.66, 2.15] & $<0.001$ & 0.015 \\
16 & 5926 & +3.25 & [2.15, 4.36] & $<0.001$ & 0.004 \\
128 & 5926 & +16.51 & [12.82, 20.20] & $<0.001$ & $<0.001$ \\
$\infty$ & 5926 & +79.68 & [74.25, 85.10] & $<0.001$ & $<0.001$ \\
\bottomrule
\end{tabular*}

\end{table}
For $L=1$, the trajectory-weighted gain is $0.01407$ (95\% CI $[0.00662,0.02153]$, $p=0.00047$), and the problem-weighted gain is $0.01220$ ($p=0.0152$). As a weighting sensitivity check, averaging trajectory means equally within each problem gives $0.01407$ ($p=0.0207$).

\paragraph{Teacher Argmax versus Lower-Ranked Replacements.}
The original A100 runs show no detectable difference between teacher-argmax and lower-ranked replacements ($p=0.27$).

\subsection{Gains by Fork Group}\label{app:flipgroups}
The group comparisons distinguish whether teacher intervention helps from whether a group benefits more than other forks. A positive gain over resampling answers the first question, while the group-versus-rest contrast answers the second (\cref{tab:flipfullsignificance}). Some contrasts pass the trajectory-level test but not the problem-level test, so their significance depends on the aggregation and weighting. For longer segments, the outcome includes additional teacher tokens and cannot be attributed to the first replacement alone (\cref{tab:flipgroups,tab:fliplongtests}).

\begin{table*}[!tbp]
\centering\small
\caption{Trajectory-weighted gains over resampling by fork group and segment length, in pp. $n_{L=1}$ counts eligible changed-token positions. Asterisks compare each group with remaining positions, not with zero, using the trajectory test.}
\label{tab:flipgroups}
\begin{tabular*}{\linewidth}{@{\extracolsep{\fill}}lrrrrr@{}}
\toprule
Group & $n_{L=1}$ & $L=1$ & $L=16$ & $L=128$ & $L=\infty$ \\
\midrule
$0.00\leq\mu_t^{T}<0.05$ & 2774 & +2.05 & +3.84 & +17.17 & +78.67\rlap{$^{*}$} \\
$0.05\leq\mu_t^{T}<0.10$ & 733 & +0.60 & +2.22 & +14.42\rlap{$^{*}$} & +81.24 \\
$0.10\leq\mu_t^{T}<0.20$ & 1007 & +0.03\rlap{$^{*}$} & +2.79 & +17.63 & +80.58 \\
$0.20\leq\mu_t^{T}<0.30$ & 706 & +1.55 & +2.79 & +16.66 & +80.83 \\
$0.30\leq\mu_t^{T}<0.50$ & 680 & +0.93 & +3.02 & +14.16\rlap{$^{*}$} & +79.19 \\
\midrule
$0.00\leq\mu_t^{S}<0.05$ & 3179 & +2.06\rlap{$^{***}$} & +3.57 & +16.85 & +78.97 \\
$0.05\leq\mu_t^{S}<0.10$ & 766 & +0.70 & +2.76 & +17.59 & +80.60 \\
$0.10\leq\mu_t^{S}<0.20$ & 885 & +0.87 & +2.98 & +15.72 & +80.74 \\
$0.20\leq\mu_t^{S}<0.30$ & 601 & +0.75 & +3.25 & +14.69 & +81.24 \\
$0.30\leq\mu_t^{S}<0.50$ & 470 & +0.06\rlap{$^{*}$} & +2.50 & +15.14 & +79.85 \\
\midrule
Blind & 531 & +5.27\rlap{$^{***}$} & +4.95\rlap{$^{*}$} & +20.70\rlap{$^{**}$} & +77.62 \\
Not blind & 5370 & +1.08\rlap{$^{***}$} & +3.00\rlap{$^{*}$} & +16.22\rlap{$^{**}$} & +79.92 \\
\midrule
Position 0.0--0.2 & 1064 & +0.74 & +2.91 & +13.56 & +80.99 \\
Position 0.2--0.4 & 1081 & +0.58\rlap{$^{*}$} & +1.27\rlap{$^{***}$} & +13.64 & +82.61\rlap{$^{*}$} \\
Position 0.4--0.6 & 1184 & +2.63 & +4.17 & +17.92 & +83.28\rlap{$^{*}$} \\
Position 0.6--0.8 & 1187 & +0.05 & +2.37 & +16.25 & +82.75\rlap{$^{*}$} \\
Position 0.8--1.0 & 1385 & +2.56 & +5.00 & +16.76 & +66.99\rlap{$^{***}$} \\
\midrule
Other word & 4003 & +1.06\rlap{$^{*}$} & +2.88 & +16.25 & +80.76\rlap{$^{**}$} \\
Reflection / contrast & 388 & +6.59\rlap{$^{**}$} & +8.88 & +21.82 & +80.33 \\
Connective & 169 & +3.81 & +5.08 & +17.76 & +78.97 \\
Digit & 254 & +0.38 & +3.93 & +14.91 & +76.89 \\
Math symbol & 752 & +1.65 & +3.37 & +17.21 & +80.08 \\
Structure / formatting & 335 & +0.86 & +1.71 & +14.00\rlap{$^{*}$} & +70.06\rlap{$^{***}$} \\
\bottomrule
\end{tabular*}

\end{table*}

\begin{table*}[!tbp]
\centering\small
\caption{Unadjusted group-versus-rest $p$-values for longer teacher segments. Superscripts give segment length. Subscripts give trajectory ($T$) or problem ($P$) aggregation.}
\label{tab:fliplongtests}
\begin{tabular*}{\linewidth}{@{\extracolsep{\fill}}lrrrrrr@{}}
\toprule
Group & $p_T^{16}$ & $p_P^{16}$ & $p_T^{128}$ & $p_P^{128}$ & $p_T^{\infty}$ & $p_P^{\infty}$ \\
\midrule
$0.00\leq\mu_t^{T}<0.05$ & 0.198 & 0.262 & 0.058 & 0.006 & 0.029 & 0.202 \\
$0.05\leq\mu_t^{T}<0.10$ & 0.218 & 0.266 & 0.049 & 0.048 & 0.079 & 0.356 \\
$0.10\leq\mu_t^{T}<0.20$ & 0.368 & 0.693 & 0.186 & 0.336 & 0.237 & 0.029 \\
$0.20\leq\mu_t^{T}<0.30$ & 0.497 & 0.489 & 0.900 & 0.908 & 0.330 & 0.422 \\
$0.30\leq\mu_t^{T}<0.50$ & 0.723 & 0.334 & 0.030 & 0.048 & 0.655 & 0.828 \\
\midrule
$0.00\leq\mu_t^{S}<0.05$ & 0.301 & 0.456 & 0.270 & 0.141 & 0.055 & 0.108 \\
$0.05\leq\mu_t^{S}<0.10$ & 0.602 & 0.509 & 0.301 & 0.610 & 0.557 & 0.389 \\
$0.10\leq\mu_t^{S}<0.20$ & 0.519 & 0.238 & 0.329 & 0.233 & 0.224 & 0.291 \\
$0.20\leq\mu_t^{S}<0.30$ & 0.921 & 0.400 & 0.186 & 0.228 & 0.082 & 0.220 \\
$0.30\leq\mu_t^{S}<0.50$ & 0.385 & 0.437 & 0.401 & 0.322 & 0.774 & 0.793 \\
\midrule
Blind & 0.013 & 0.442 & 0.003 & 0.067 & 0.244 & 0.072 \\
Not blind & 0.013 & 0.442 & 0.003 & 0.067 & 0.244 & 0.072 \\
\midrule
Position 0.0--0.2 & 0.612 & 0.751 & 0.123 & 0.394 & 0.386 & 0.228 \\
Position 0.2--0.4 & $<0.001$ & 0.010 & 0.177 & 0.605 & 0.024 & 0.165 \\
Position 0.4--0.6 & 0.358 & 0.058 & 0.231 & 0.208 & 0.015 & 0.104 \\
Position 0.6--0.8 & 0.483 & 0.462 & 0.881 & 0.774 & 0.044 & 0.159 \\
Position 0.8--1.0 & 0.118 & 0.470 & 0.824 & 0.965 & $<0.001$ & 0.041 \\
\midrule
Other word & 0.384 & 0.557 & 0.884 & 0.769 & 0.003 & 0.049 \\
Reflection / contrast & 0.130 & 0.267 & 0.109 & 0.434 & 0.082 & 0.104 \\
Connective & 0.644 & 0.509 & 0.961 & 0.905 & 0.395 & 0.205 \\
Digit & 0.909 & 0.474 & 0.561 & 0.275 & 0.363 & 0.160 \\
Math symbol & 0.945 & 0.975 & 0.520 & 0.713 & 0.856 & 0.725 \\
Structure / formatting & 0.193 & 0.717 & 0.025 & 0.256 & $<0.001$ & 0.012 \\
\bottomrule
\end{tabular*}

\end{table*}

\subsection{Group Eligibility and Success Rates}\label{app:fliprates}
Larger intervention gains do not necessarily identify positions with the highest recovery rates. Blind forks have larger gains than non-blind forks, but their replacement success rates are much closer than their resampling success rates (\cref{tab:fliprates}). Their larger gain mainly reflects the lower resampling success at these positions. The two tables also use different eligibility rules. \Cref{tab:flipfullsignificance} estimates within-group gains using all units with eligible positions, while \cref{tab:fliprates} retains trajectories with at least three eligible positions in both the group and its complement. This changes the positions and trajectory weights contributing to the estimates, including the reported reflection gain.
\begin{table*}[!tbp]
\centering\footnotesize\setlength{\tabcolsep}{3pt}
\caption{Complete single-flip comparisons, with effects in pp and shares of 5{,}901 eligible forks. Within-group gains use units with eligible positions. Group-versus-rest tests require at least three positions per side. $n_T/n_P$ counts eligible trajectories/problems for these contrasts. Tests are unadjusted (\cref{app:stats_singleflip}).}
\label{tab:flipfullsignificance}
\begin{tabular*}{\textwidth}{@{\extracolsep{\fill}}lrrrrrrrrrr@{}}
\toprule
 & & \multicolumn{4}{c}{Flip $-$ Resample} & \multicolumn{4}{c}{Net Gain $-$ Rest} &  \\
\cmidrule(lr){3-6}\cmidrule(lr){7-10}
 & & \multicolumn{2}{c}{Trajectory} & \multicolumn{2}{c}{Problem} & \multicolumn{2}{c}{Trajectory} & \multicolumn{2}{c}{Problem} &  \\
Fork Group & Share (\%) & Gain & $p_T$ & Gain & $p_P$ & $\Delta$ & $p_T$ & $\Delta$ & $p_P$ & $n_T/n_P$ \\
\midrule
All Forks & 100.0 & +1.41\makebox[1.2em][l]{$^{***}$} & $<0.001$ & +1.22\makebox[1.2em][l]{$^{*}$} & 0.015 & -- & -- & -- & -- & -- \\
\midrule
$0.00\leq\mu_t^{T}<0.05$ & 47.0 & +2.05\makebox[1.2em][l]{$^{***}$} & $<0.001$ & +1.94\makebox[1.2em][l]{$^{*}$} & 0.032 & +1.17\makebox[1.2em][l]{} & 0.057 & +1.25\makebox[1.2em][l]{} & 0.145 & 40/10 \\
$0.05\leq\mu_t^{T}<0.10$ & 12.4 & +0.60\makebox[1.2em][l]{} & 0.133 & +0.69\makebox[1.2em][l]{} & 0.059 & -0.88\makebox[1.2em][l]{} & 0.102 & -0.60\makebox[1.2em][l]{} & 0.244 & 40/10 \\
$0.10\leq\mu_t^{T}<0.20$ & 17.1 & +0.03\makebox[1.2em][l]{} & 0.961 & +0.04\makebox[1.2em][l]{} & 0.941 & -1.56\makebox[1.2em][l]{$^{*}$} & 0.021 & -1.37\makebox[1.2em][l]{} & 0.106 & 40/10 \\
$0.20\leq\mu_t^{T}<0.30$ & 12.0 & +1.55\makebox[1.2em][l]{$^{*}$} & 0.044 & +1.33\makebox[1.2em][l]{} & 0.160 & +0.13\makebox[1.2em][l]{} & 0.848 & +0.11\makebox[1.2em][l]{} & 0.883 & 40/10 \\
$0.30\leq\mu_t^{T}<0.50$ & 11.5 & +0.93\makebox[1.2em][l]{} & 0.302 & +0.76\makebox[1.2em][l]{} & 0.509 & -0.59\makebox[1.2em][l]{} & 0.551 & -0.60\makebox[1.2em][l]{} & 0.665 & 40/10 \\
\midrule
$0.00\leq\mu_t^{S}<0.05$ & 53.9 & +2.06\makebox[1.2em][l]{$^{***}$} & $<0.001$ & +1.83\makebox[1.2em][l]{$^{**}$} & 0.007 & +1.40\makebox[1.2em][l]{$^{***}$} & $<0.001$ & +1.35\makebox[1.2em][l]{$^{**}$} & 0.004 & 40/10 \\
$0.05\leq\mu_t^{S}<0.10$ & 13.0 & +0.70\makebox[1.2em][l]{} & 0.198 & +0.66\makebox[1.2em][l]{} & 0.141 & -0.57\makebox[1.2em][l]{} & 0.224 & -0.64\makebox[1.2em][l]{} & 0.257 & 39/10 \\
$0.10\leq\mu_t^{S}<0.20$ & 15.0 & +0.87\makebox[1.2em][l]{} & 0.109 & +0.56\makebox[1.2em][l]{} & 0.371 & -0.62\makebox[1.2em][l]{} & 0.214 & -0.78\makebox[1.2em][l]{} & 0.247 & 40/10 \\
$0.20\leq\mu_t^{S}<0.30$ & 10.2 & +0.75\makebox[1.2em][l]{} & 0.327 & +0.55\makebox[1.2em][l]{} & 0.479 & -1.25\makebox[1.2em][l]{} & 0.125 & -0.76\makebox[1.2em][l]{} & 0.303 & 36/10 \\
$0.30\leq\mu_t^{S}<0.50$ & 8.0 & +0.06\makebox[1.2em][l]{} & 0.922 & +0.15\makebox[1.2em][l]{} & 0.829 & -1.46\makebox[1.2em][l]{$^{*}$} & 0.049 & -1.17\makebox[1.2em][l]{} & 0.053 & 38/10 \\
\midrule
Blind & 9.0 & +5.27\makebox[1.2em][l]{$^{***}$} & $<0.001$ & +4.27\makebox[1.2em][l]{$^{*}$} & 0.012 & +4.08\makebox[1.2em][l]{$^{***}$} & $<0.001$ & +3.31\makebox[1.2em][l]{$^{*}$} & 0.015 & 36/10 \\
Not Blind & 91.0 & +1.08\makebox[1.2em][l]{$^{**}$} & 0.002 & +0.96\makebox[1.2em][l]{$^{*}$} & 0.032 & -4.08\makebox[1.2em][l]{$^{***}$} & $<0.001$ & -3.31\makebox[1.2em][l]{$^{*}$} & 0.015 & 36/10 \\
\midrule
Position 0.0--0.2 & 18.0 & +0.74\makebox[1.2em][l]{} & 0.302 & +0.74\makebox[1.2em][l]{} & 0.315 & -0.91\makebox[1.2em][l]{} & 0.295 & -0.73\makebox[1.2em][l]{} & 0.493 & 40/10 \\
Position 0.2--0.4 & 18.3 & +0.58\makebox[1.2em][l]{} & 0.209 & +0.65\makebox[1.2em][l]{} & 0.366 & -1.04\makebox[1.2em][l]{$^{*}$} & 0.040 & -0.75\makebox[1.2em][l]{} & 0.215 & 40/10 \\
Position 0.4--0.6 & 20.1 & +2.63\makebox[1.2em][l]{$^{**}$} & 0.006 & +2.50\makebox[1.2em][l]{$^{**}$} & 0.006 & +1.46\makebox[1.2em][l]{} & 0.084 & +1.53\makebox[1.2em][l]{$^{**}$} & 0.009 & 39/10 \\
Position 0.6--0.8 & 20.1 & +0.05\makebox[1.2em][l]{} & 0.959 & +0.10\makebox[1.2em][l]{} & 0.905 & -1.45\makebox[1.2em][l]{} & 0.234 & -1.24\makebox[1.2em][l]{} & 0.300 & 40/10 \\
Position 0.8--1.0 & 23.5 & +2.56\makebox[1.2em][l]{$^{**}$} & 0.005 & +2.09\makebox[1.2em][l]{} & 0.105 & +1.49\makebox[1.2em][l]{} & 0.079 & +1.12\makebox[1.2em][l]{} & 0.344 & 40/10 \\
\midrule
Other Word & 67.8 & +1.06\makebox[1.2em][l]{$^{**}$} & 0.002 & +0.92\makebox[1.2em][l]{$^{*}$} & 0.023 & -0.91\makebox[1.2em][l]{$^{*}$} & 0.026 & -0.77\makebox[1.2em][l]{} & 0.052 & 40/10 \\
Reflection / Contrast & 6.6 & +6.59\makebox[1.2em][l]{$^{*}$} & 0.045 & +5.11\makebox[1.2em][l]{} & 0.116 & +1.84\makebox[1.2em][l]{$^{**}$} & 0.008 & +3.84\makebox[1.2em][l]{} & 0.189 & 30/9 \\
Connective & 2.9 & +3.81\makebox[1.2em][l]{} & 0.117 & +1.85\makebox[1.2em][l]{} & 0.198 & +0.58\makebox[1.2em][l]{} & 0.755 & +0.56\makebox[1.2em][l]{} & 0.729 & 22/8 \\
Digit & 4.3 & +0.38\makebox[1.2em][l]{} & 0.695 & +1.91\makebox[1.2em][l]{} & 0.064 & +1.28\makebox[1.2em][l]{} & 0.123 & +0.75\makebox[1.2em][l]{} & 0.319 & 28/10 \\
Math Symbol & 12.7 & +1.65\makebox[1.2em][l]{} & 0.145 & +1.35\makebox[1.2em][l]{} & 0.096 & +0.26\makebox[1.2em][l]{} & 0.811 & +0.14\makebox[1.2em][l]{} & 0.846 & 39/10 \\
Structure / Formatting & 5.7 & +0.86\makebox[1.2em][l]{} & 0.358 & +0.61\makebox[1.2em][l]{} & 0.177 & -1.18\makebox[1.2em][l]{} & 0.229 & -0.68\makebox[1.2em][l]{} & 0.079 & 36/10 \\
\bottomrule
\end{tabular*}

\end{table*}

\begin{table*}[!tbp]
\centering\small
\caption{Single-token replacement and resampling success rates. Gain is flip minus resample success at the same positions. Each group pair uses trajectories with at least three eligible positions per side. Rates, gains, and 95\% intervals use equal trajectory weights. Group contrasts are reported in \cref{tab:flipmain,tab:flipfullsignificance}.}
\label{tab:fliprates}
\begin{tabular*}{\linewidth}{@{\extracolsep{\fill}}lrrrr@{}}
\toprule
Fork group & $n$ & Flip (\%) & Resample (\%) & Gain [95\% CI] (pp) \\
\midrule
All eligible forks & 5,901 & 11.31 & 9.91 & +1.41 [0.66, 2.15] \\
\midrule
$\mu_t^S<0.05$ & 3,179 & 11.22 & 9.16 & +2.06 [1.13, 2.99] \\
$\mu_t^S\geq0.05$ & 2,722 & 11.50 & 10.84 & +0.66 [0.02, 1.30] \\
\midrule
Blind & 524 & 10.19 & 5.35 & +4.84 [2.29, 7.39] \\
Not blind & 5,146 & 9.56 & 8.80 & +0.77 [0.19, 1.34] \\
\midrule
Reflection/contrast & 383 & 8.57 & 5.79 & +2.78 [1.15, 4.41] \\
Other tokens & 4,760 & 9.43 & 8.49 & +0.94 [0.34, 1.54] \\
\bottomrule
\end{tabular*}

\end{table*}

\subsection{Predicting Single-Flip Gains over Resampling}\label{app:prediction}
For prediction at $L=1$, the label is whether estimated true-flip success exceeds the independently estimated resample rate at the same fork. Of 5{,}901 eligible positions, 3{,}303 are tied and excluded, leaving 2{,}598 labels with 1{,}333 positives. These labels concern individual forks rather than group-mean gains.

Logistic regression uses five-fold cross-validation grouped by trajectory, with numeric and one-hot token-class features standardized within each training fold. The 95\% intervals use 100 trajectory-bootstrap refits, keeping duplicate copies of a trajectory in the same fold. We report AUC only at $L=1$ because longer teacher continuations do not isolate the initial token's predictive value.

\begin{table}[!htbp]
\centering\footnotesize\setlength{\tabcolsep}{3pt}
\caption{Cross-validated AUC for predicting whether a single flip beats resampling, on 2{,}598 non-tied forks. Intervals use trajectory-bootstrap refits.}
\label{tab:flipauc}
\begin{tabular*}{\linewidth}{@{\extracolsep{\fill}}llr@{}}
\toprule
Group & Feature & AUC [95\% CI] \\
\midrule
Token mass & $\mu_t^T$ & 0.512 [0.453, 0.537] \\
 & $\mu_t^S$ & 0.539 [0.503, 0.557] \\
Confidence & student & 0.480 [0.435, 0.510] \\
 & teacher & 0.488 [0.434, 0.517] \\
Entropy & student & 0.486 [0.440, 0.510] \\
 & teacher & 0.498 [0.444, 0.519] \\
Divergence & reverse KL & 0.532 [0.491, 0.553] \\
 & forward KL & 0.548 [0.505, 0.565] \\
Token identity & blind spot & 0.499 [0.467, 0.513] \\
 & reflection word & 0.495 [0.460, 0.504] \\
 & token class & 0.503 [0.462, 0.519] \\
Position & position & 0.522 [0.463, 0.555] \\
 & fork score & 0.472 [0.427, 0.519] \\
Combined & distributional & 0.554 [0.505, 0.568] \\
 & position + score & 0.516 [0.451, 0.546] \\
 & all thirteen & 0.554 [0.503, 0.567] \\
\bottomrule
\end{tabular*}

\end{table}

\subsection{Keep-Token Baseline}\label{app:keep}
The keep arm preserves the original student token at each of the 5{,}926 fork prefixes and independently samples 16 continuations, giving 94{,}816 outcomes. It uses the same L40S continuation pipeline and decoding settings as the replacement and resample arms. True-flip comparisons retain the same 5{,}901 eligible positions as the main analysis. Resample versus keep uses all 5{,}926 positions.

\begin{table*}[!tbp]
\centering\small
\caption{Keep-token baseline checks. Gains and Student-$t$ 95\% intervals are trajectory-weighted. $p_T$ and $p_P$ use trajectory and problem units, respectively, as in \cref{app:stats_singleflip}.}
\label{tab:keep}
\begin{tabular*}{\linewidth}{@{\extracolsep{\fill}}lrrr@{}}
\toprule
Comparison & Gain [95\% CI] (pp) & $p_T$ & $p_P$ \\
\midrule
True flip $-$ keep & +1.55 [0.79, 2.32] & 0.00020 & 0.01697 \\
Resample $-$ keep & +0.14 [-0.22, 0.50] & 0.43082 & 0.31733 \\
\bottomrule
\end{tabular*}

\end{table*}

Replacement improves over keep under both aggregation-level tests. Resample and keep do not differ detectably (\cref{tab:keep}).

\appendixsection{Intervention Timing and Recovery}\label{app:singleposition}
The success rates below average positions, whereas paired gains average within trajectories before averaging across trajectories. Subtracting the displayed success rates therefore need not reproduce the trajectory-weighted gains.
\begin{figure}[!htbp]
\centering
\includegraphics[width=\linewidth,height=0.32\textheight,keepaspectratio]{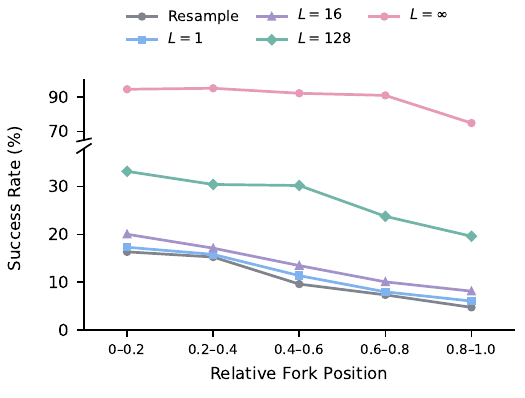}
\caption{Success by fork position and teacher-segment length, averaged over positions in each fifth. The vertical axis omits 38--65\%. $L=1$ retains changed-token samples. Longer segments use the original continuation runs (\cref{app:compute}).}
\label{fig:positionsuccess}
\end{figure}
\begin{figure*}[!tbp]
\centering
\includegraphics[width=0.90\linewidth]{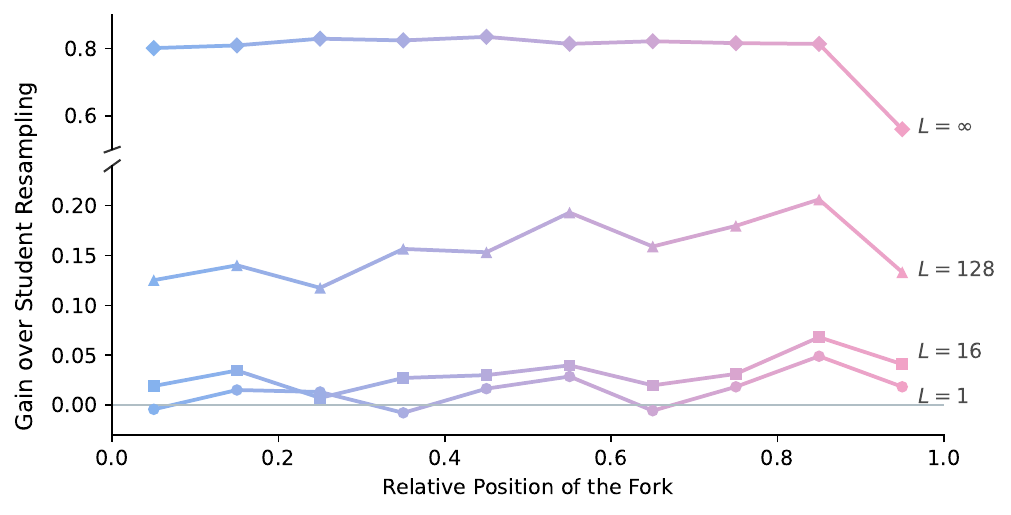}
\caption{Gain over the complete resample baseline by position decile and teacher-segment length. Paired position differences are averaged within each trajectory and then across trajectories. At $L=1$ only paths changing the original token are included.}
\label{fig:flippos}
\end{figure*}

\begin{table*}[!tbp]
\centering\small\setlength{\tabcolsep}{4pt}
\caption{Position-averaged success rates by trajectory fifth. Resample uses all positions, and $L=1$ uses positions with an actual token change. Paired gains and group tests are reported separately in \cref{tab:flipgroups,tab:flipfullsignificance}.}
\label{tab:position}
\begin{tabular*}{\linewidth}{@{\extracolsep{\fill}}lrrrrr@{}}
\toprule
Position & Resample & $L=1$ & $L=16$ & $L=128$ & $L=\infty$ \\
\midrule
All & 0.103 & 0.113 & 0.134 & 0.270 & 0.887 \\
\midrule
0.0--0.2 & 0.163 & 0.173 & 0.200 & 0.332 & 0.944 \\
0.2--0.4 & 0.153 & 0.158 & 0.171 & 0.305 & 0.949 \\
0.4--0.6 & 0.096 & 0.114 & 0.135 & 0.302 & 0.920 \\
0.6--0.8 & 0.073 & 0.080 & 0.101 & 0.238 & 0.908 \\
0.8--1.0 & 0.047 & 0.060 & 0.081 & 0.196 & 0.748 \\
\bottomrule
\end{tabular*}

\end{table*}

Absolute recovery declines at later positions under both teacher intervention and student resampling (\cref{tab:position}). After accounting for resampling, no single-token position group differs from the remaining forks under both trajectory- and problem-level tests. Position therefore does not reliably distinguish which single-token interventions provide larger gains. Across all tested intervention lengths, however, gains fall from the penultimate to the final decile (\cref{fig:flippos}). This drop is largest when the teacher completes the continuation, from 81.4 to 56.1 pp. Longer interventions measure the continuation's contribution rather than that of the initial token.

\appendixsection{Student Ability to Follow Teacher Corrections}\label{app:continuation}
\subsection{Fork Spacing on Teacher Trajectories}\label{app:teacherspacing}
We score 4{,}000 teacher-generated trajectories with the student and teacher, identifying forks at 2.11M of 15.17M positions. Fork spacing is the position difference between successive forks within the same trajectory. A consecutive run is a maximal sequence of adjacent forks.

\begin{figure*}[!tbp]
\centering
\includegraphics[width=0.90\linewidth]{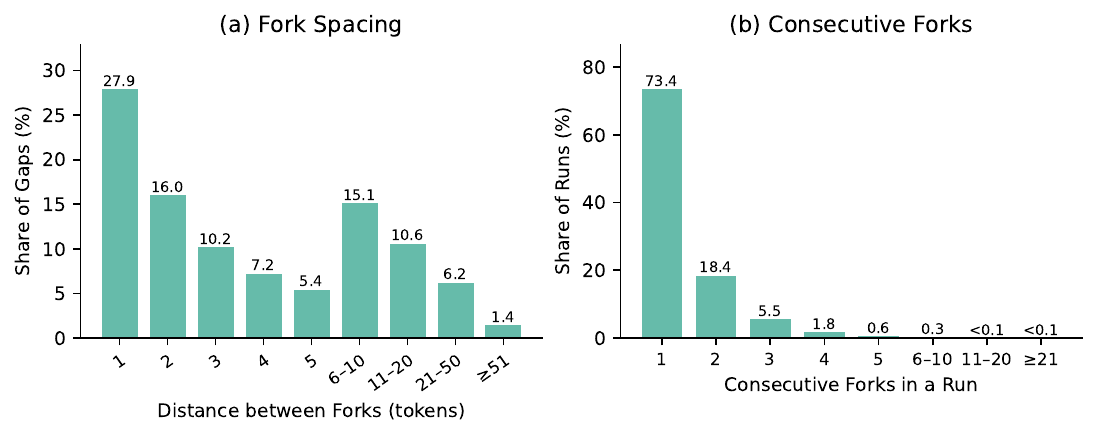}
\caption{Fork spacing on teacher-generated trajectories. (a) Within-trajectory distances between successive forks. (b) Lengths of maximal consecutive fork runs. Percentages use gaps and runs as their respective denominators.}
\label{fig:teacherspacing}
\end{figure*}

The median distance between successive forks is three tokens, and 27.9\% occur at adjacent positions (\cref{fig:teacherspacing}). Although 47.0\% of forks belong to runs of at least two tokens, most runs contain only one or two tokens. These short runs suggest that disagreement on teacher trajectories is usually local rather than sustained over long stretches.

\subsection{Student Responses after a Teacher Correction}\label{app:afterflipdiagnostics}
We examine difficulty following teacher corrections through student entropy and teacher-token probability. The unchanged-token control retains teacher samples reproducing the original student token at 1{,}495 positions, with 1{,}470 admitting both changed and unchanged paths. This control is used only for these diagnostics, not as the independent resample baseline.
\begin{figure*}[!tbp]
\centering
\includegraphics[width=0.90\linewidth,height=0.55\textheight,keepaspectratio]{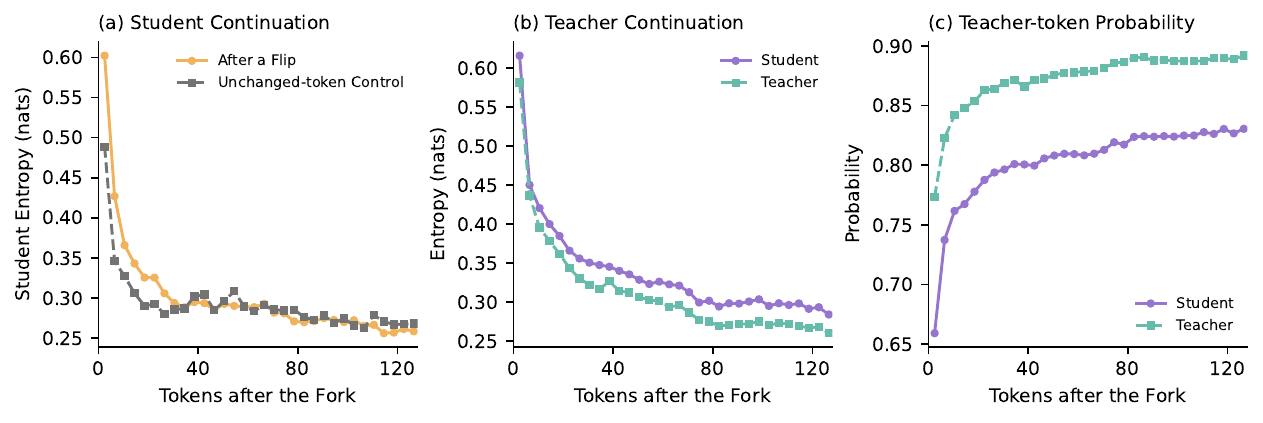}
\caption{Token-level diagnostics after a single flip. Each point averages four offsets, with up to 5{,}901 changed-token forks and 1{,}495 unchanged-token controls. (a) Student entropy after changed and unchanged tokens. (b) Both models' entropy on the teacher's continuation. (c) Their probabilities of the teacher's tokens.}
\label{fig:afterflip}
\end{figure*}
Student entropy is elevated for about 40 tokens after a replacement before approaching the unchanged-token control (\cref{fig:afterflip}a). On the teacher's continuation, student entropy tracks teacher entropy within $0.05$ nats (b). The probability the student assigns to teacher tokens rises from $0.61$ to $0.83$, remaining $0.06$ to $0.13$ below the teacher's own probability (c).

\appendixsection{Additional Dual-Track Results}\label{app:dualresults}
\subsection{Complete Results for the Main Model Pair}\label{app:dualmainresults}\label{app:dualcontrols}
Random controls use preset selection probabilities chosen to approximate trigger coverage. We report realized flip counts because each policy changes its subsequent trajectory, so coverage matching does not imply equal flip counts.

\begin{table*}[!tbp]
\centering\small\setlength{\tabcolsep}{3pt}
\caption{Selected dual-track controls. F/T is mean flips per trajectory and F/F is the percentage of forks flipped. Test details are in \cref{app:stats_dualtrack}.}
\label{tab:dual}
\begin{tabular*}{\columnwidth}{@{\extracolsep{\fill}}lrrr@{}}
\toprule
Trigger & Acc. (\%) & F/T & F/F (\%) \\
\midrule
$\mu^{T}_t<0.1$\rlap{$^{\boldsymbol{***}}$} & 71.0 & 199 & 46 \\
\textcolor{black!60}{Random ($p=0.46$)} & \textcolor{black!60}{57.3} & \textcolor{black!60}{157} & \textcolor{black!60}{45} \\
\midrule
$\mu^{T}_t<0.05$\rlap{$^{\boldsymbol{***}}$} & 65.2 & 143 & 35 \\
\textcolor{black!60}{Random ($p=0.35$)} & \textcolor{black!60}{52.0} & \textcolor{black!60}{103} & \textcolor{black!60}{34} \\
\midrule
$\mu^{T}_t<10^{-5}$\rlap{$^{\boldsymbol{**}}$} & 38.4 & 4 & 2 \\
\textcolor{black!60}{Random ($p=0.06$)} & \textcolor{black!60}{32.9} & \textcolor{black!60}{11} & \textcolor{black!60}{6} \\
\bottomrule
\end{tabular*}

\end{table*}

\begin{table}[!htbp]
\centering\footnotesize\setlength{\tabcolsep}{3pt}
\caption{All points in \cref{fig:flipscaling}. Each uses 200 problems with 4 trajectories each. F/T counts actual replacements per trajectory. Random controls are shaded.}
\label{tab:dualcomplete}
\begin{tabular*}{\columnwidth}{@{\extracolsep{\fill}}lrr@{}}
\toprule
Trigger & Acc. (\%) & F/T \\
\midrule
$\mu_t^T<10^{-5}$ & 38.38 & 4.11 \\
$\mu_t^T<10^{-4}$ & 42.88 & 10.27 \\
$\mu_t^T<10^{-3}$ & 51.50 & 25.75 \\
$\mu_t^T<0.05$ & 65.25 & 142.50 \\
$\mu_t^T<0.1$ & 71.00 & 199.01 \\
\midrule
$\mathrm{KL}>10$ & 35.12 & 3.06 \\
$\mathrm{KL}>8$ & 41.88 & 6.64 \\
$\mathrm{KL}>5$ & 54.25 & 23.66 \\
$\mathrm{KL}>4$ & 55.38 & 41.25 \\
$\mathrm{KL}>3$ & 63.50 & 77.43 \\
$\mathrm{KL}>2$ & 71.12 & 152.46 \\
$\mathrm{KL}>1$ & 73.75 & 311.64 \\
\midrule
\rowcolor{black!5}
Random ($p=0.06$) & 32.88 & 11.13 \\
\rowcolor{black!5}
Random ($p=0.11$) & 36.12 & 22.82 \\
\rowcolor{black!5}
Random ($p=0.22$) & 42.62 & 52.76 \\
\rowcolor{black!5}
Random ($p=0.35$) & 52.00 & 102.62 \\
\rowcolor{black!5}
Random ($p=0.46$) & 57.25 & 156.88 \\
\rowcolor{black!5}
Random ($p=0.90$) & 74.50 & 407.83 \\
\midrule
Reflection & 34.88 & 13.72 \\
\bottomrule
\end{tabular*}

\end{table}

The original controls give accuracy differences of $+0.138$ ($[0.099,0.176]$, $p<10^{-10}$) for $\mu_t^T<0.1$ versus random 46\%, and $+0.133$ ($[0.093,0.173]$, $p<10^{-9}$) for $\mu_t^T<0.05$ versus random 35\%. Reflection-word selection versus random 6\% gives $+0.020$ ($[-0.017,0.057]$, $p=0.29$). The sparse $\mu_t^T<10^{-5}$ rule gives $+0.055$ ($[0.018,0.092]$, $p=0.004$) versus random 6\% and $+0.035$ ($[-0.003,0.073]$, $p=0.07$) versus reflection-word selection.

Additional random controls use probabilities 0.11 and 0.22. Against 0.11, $\mu_t^T<10^{-3}$ gains 15.375 pp ($p=1.18\times10^{-10}$), and $\mathrm{KL}>5$ gains 18.125 pp ($p=8.02\times10^{-14}$). Against 0.22, $\mathrm{KL}>4$ gains 12.75 pp ($p=1.36\times10^{-9}$). All comparisons use the same problem-paired test.

\subsection{Handoff}\label{app:handoff}
We are concerned that the accuracy gains may come from the teacher supplying the answer near the end of the trajectory. We therefore stop teacher intervention at earlier fractions $f$ of each dual-track trajectory before the answer zone and let the student complete the reasoning and answer alone with no further flips (\cref{app:handoff_settings}). Varying $f$ from 0 (the student alone) to 1 (the full dual-track trajectory) measures accuracy as a function of how far the teacher's corrections extend.

\begin{figure}[!htbp]
\centering
\includegraphics[width=\linewidth]{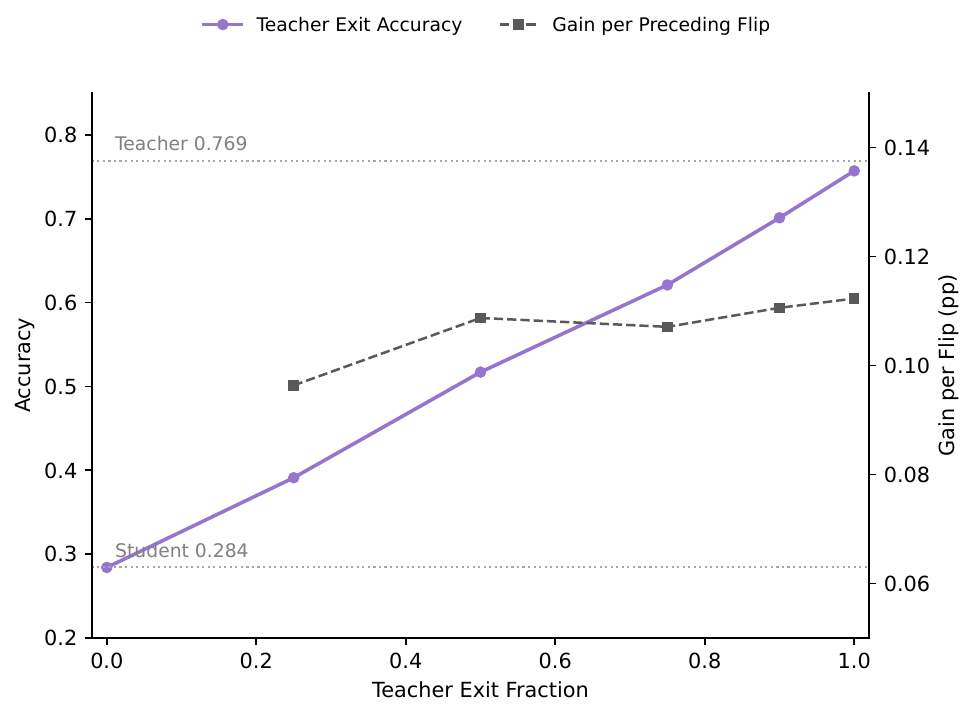}
\caption{Accuracy after teacher exit at fraction $f$ (left axis) and gain over the student alone per preceding flip, in pp (right). Dotted lines mark standalone student and teacher accuracy.}
\label{fig:handoff}
\end{figure}

\subsubsection{Gains across Teacher Exit Positions}
Accuracy gains accumulate roughly in proportion to the number of flips across teacher exit positions, with an average gain of 0.10--0.11 pp per flip (\cref{fig:handoff}). This suggests that the teacher's contribution is spread across the trajectory rather than concentrated in a narrow region.

Each dual-track trajectory is cut at $\lfloor f\cdot t_{\mathrm{ans}}\rceil$, where $t_{\mathrm{ans}}$ is the first position of the answer-zone marker. We grade the complete text after the student finishes using the settings in \cref{app:compute}. At $f=1$, the student writes only the answer line (median 10 tokens), with accuracy $0.757$ compared with dual-track accuracy $0.745$.

Under the constant-gain-per-flip model in \cref{app:stats_handoff}, mean prediction residuals range from $-0.02$ to $-0.01$. Trajectory-level tests give $p=0.06$, $0.20$, $0.06$, and $0.34$ at $f=0.25$, $0.5$, $0.75$, and $0.9$, respectively. All problem-level tests give $p\geq0.13$, and the residual-slope test gives $p=0.47$.

\subsection{Ablation across Model Sizes}\label{app:additionalpairs}
\begin{figure*}[!tbp]
\centering
\includegraphics[width=\linewidth]{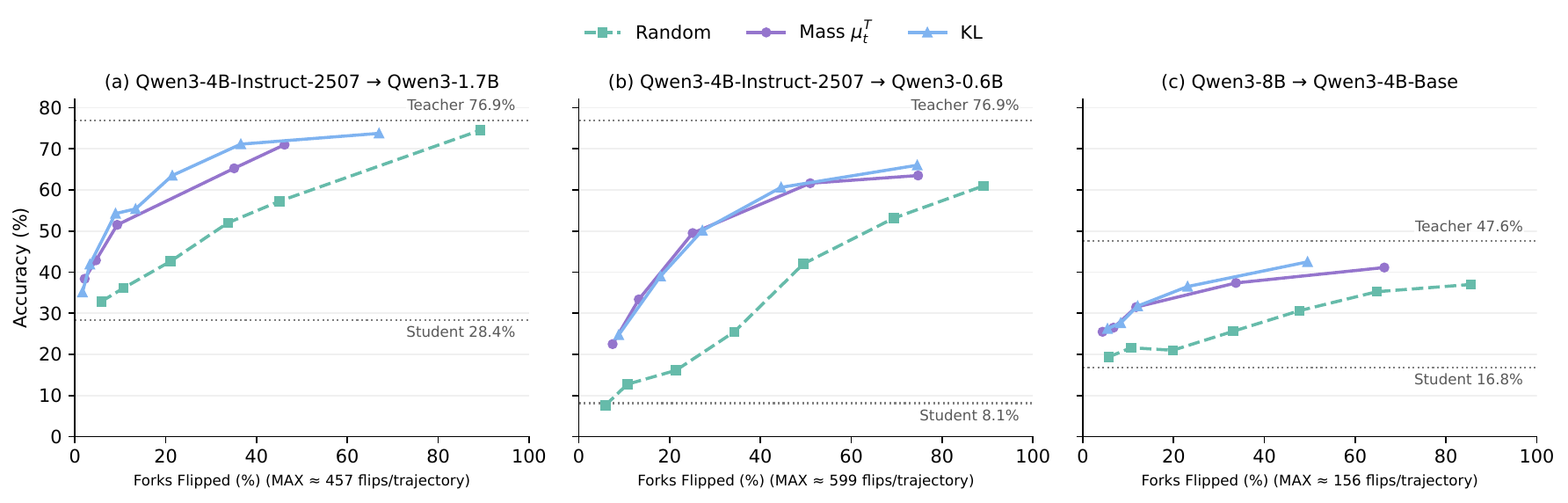}
\caption{Accuracy versus the percentage of forks flipped for three teacher--student pairs, with 200 problems and four trajectories per point. Dotted lines mark standalone baselines. MAX estimates the 100\% count using each pair's highest-coverage random run.}
\label{fig:dualthreepairs}
\end{figure*}

We test two additional teacher--student pairs within the Qwen family, Qwen3-4B-Instruct-2507 to Qwen3-0.6B and Qwen3-8B to Qwen3-4B-Base (\cref{app:additionalpair_settings}). Both pairs show the same overall pattern as the original 4B-to-1.7B pair (\cref{fig:dualthreepairs}). Mass- and KL-based selection achieve higher accuracy than random flips at intervention counts of similar magnitude, while increasing random flips also improves accuracy.

\subsection{Ablation on a Different Dataset}\label{app:ifeval}
\begin{figure}[!htbp]
\centering
\includegraphics[width=\linewidth]{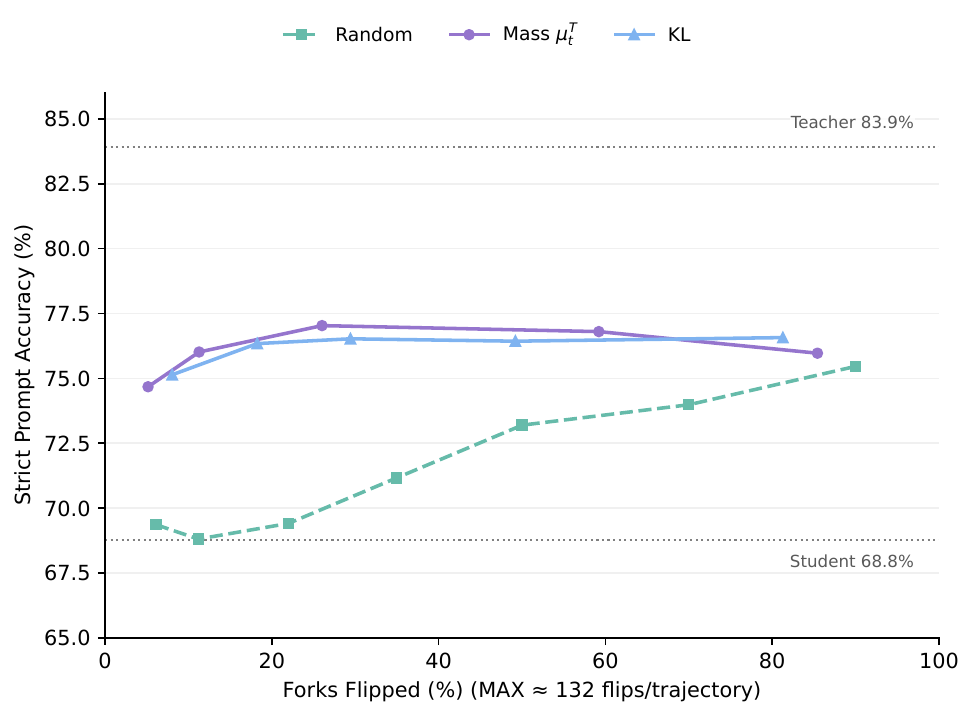}
\caption{Strict prompt accuracy on official IFEval versus the percentage of forks flipped (\cref{app:ifeval_settings}). Dotted lines mark standalone baselines. MAX estimates the 100\% count using the highest-coverage random run.}
\label{fig:ifevaldual}
\end{figure}

We repeat the interventions on official IFEval using the original Qwen teacher--student pair. Mass- and KL-based selection remain beneficial, especially at low flip budgets, while random selection gradually approaches the selective curves as coverage increases (\cref{fig:ifevaldual}).

Yet accuracy remains around 76\% even when nearly 90\% of forks are flipped, below the teacher's 83.9\%. High fork coverage does not reproduce teacher-only generation because the student still samples its own tokens wherever the models share the same argmax. The teacher may place greater probability on the correct token or assign more probability to suitable alternatives. These possibilities motivate further study of the teacher's advantage beyond argmax disagreements.

\subsection{Ablation across Model Families}\label{app:otherfamilies}
\begin{figure*}[!tbp]
\centering
\includegraphics[width=\linewidth]{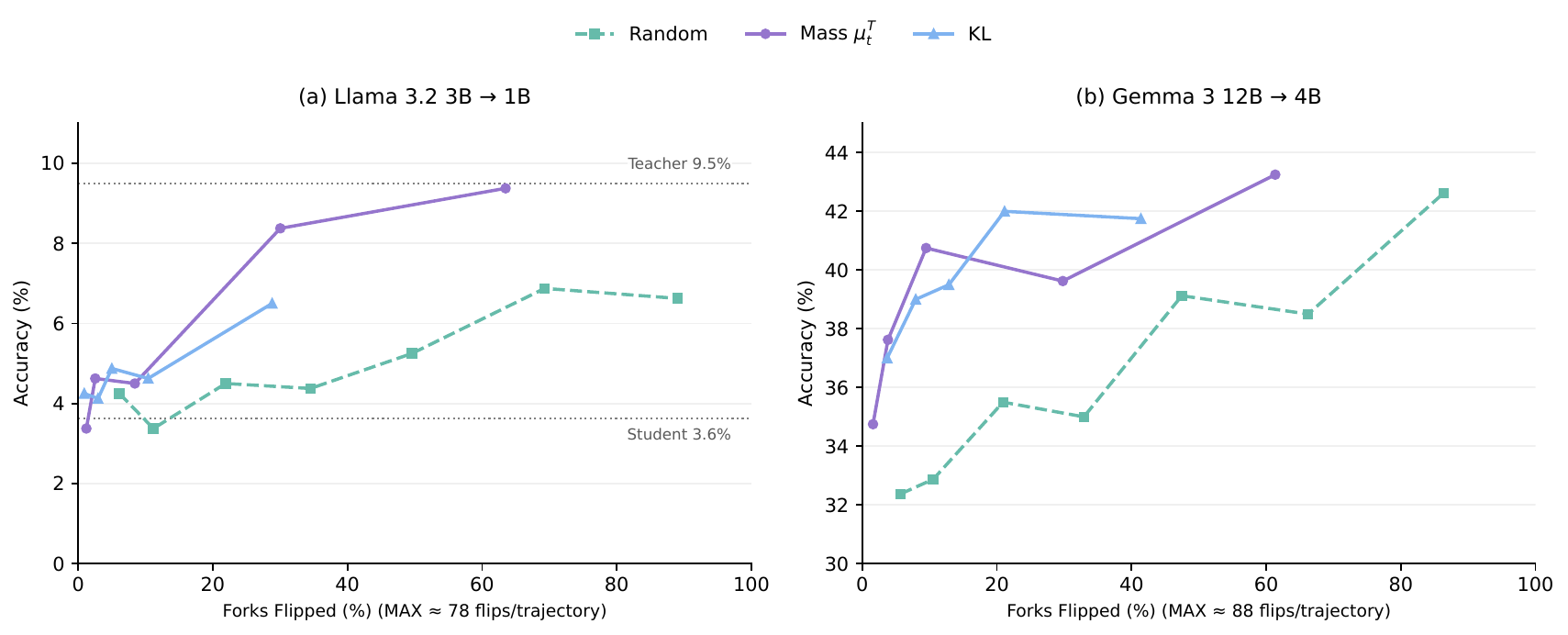}
\caption{Dual-track accuracy on DAPO-Math for Llama 3.2 3B to 1B and Gemma 3 12B to 4B (\cref{app:llama_settings,app:gemma_settings}). Each point uses 200 problems and four trajectories per problem. Dotted lines show Llama's same-problem baselines with an 8{,}192-token context limit versus 12{,}288 for interventions. MAX estimates the 100\% count using each pair's highest-coverage random run.}
\label{fig:llamadual}
\label{fig:gemmadual}
\end{figure*}

We also test Llama-3.2-3B-Instruct as teacher and Llama-3.2-1B-Instruct as student on DAPO-Math (\cref{app:llama_settings}). Mass-based selection achieves higher accuracy than random flips at similar intervention counts (\cref{fig:llamadual}). At approximately 20--25 flips per trajectory, accuracy is 8.4\% with mass-based selection, 6.5\% with KL-based selection, and 4.4\% with random selection.

The same KL threshold covers substantially fewer forks in this pair. $\mathrm{KL}>1$ selects 28.8\% of encountered Llama forks, compared with 49.5--74.5\% in the two additional Qwen pairs using the same estimator (\cref{fig:dualthreepairs,app:additionalpair_settings}). Thus, a fixed KL threshold does not provide comparable intervention coverage across model pairs, even though mass-based selection remains useful in the Llama setting.

\subsection{Results by Problem Group}\label{app:dualgroups}
Teacher intervention can also reduce accuracy on problems the student can solve but the teacher cannot (\cref{tab:cells}). In this group of 16 trajectories, accuracy falls from 31.2\% for the student alone to 18.8\% under either tested mass threshold. This small subgroup illustrates that replacing student choices with teacher choices is not uniformly beneficial, even when the teacher performs better overall.

\begin{table*}[!tbp]
\centering\small\setlength{\tabcolsep}{4pt}
\caption{Accuracy by whether the student (S) and teacher (T) solve a problem on their own (pass@4 $>0$). $n$ counts trajectories.}
\label{tab:cells}
\begin{tabular*}{\linewidth}{@{\extracolsep{\fill}}lrcccc@{}}
\toprule
 & $n$ & Student & $\tau{=}0.05$ & $\tau{=}0.1$ & Teacher \\
\midrule
S cannot, T can & 308 & 0.065 & 0.575 & 0.666 & 0.821 \\
Both can & 388 & 0.505 & 0.871 & 0.905 & 0.933 \\
Neither can & 88 & 0.034 & 0.045 & 0.102 & 0.000 \\
S can, T cannot & 16 & 0.312 & 0.188 & 0.188 & 0.000 \\
\bottomrule
\end{tabular*}

\end{table*}

Problem groups use the student and teacher's standalone pass@4 on these same 200 problems. A model counts as solving a problem if at least one of its four samples is correct.

\subsection{Teacher Sampling versus Argmax}\label{app:teachersampling}
\begin{table}[!htbp]
\centering\footnotesize\setlength{\tabcolsep}{3pt}
\caption{Teacher sampling versus argmax under the same mass trigger, on 200 problems with 4 trajectories each. Sampled replacements use temperature 0.7 and top-$p$ 0.95. Gain is sample minus argmax, and $p$ is from a two-sided test paired over problems (\cref{app:stats_dualtrack}).}
\label{tab:dualsampling}
\begin{tabular*}{\linewidth}{@{\extracolsep{\fill}}lrrrr@{}}
\toprule
Trigger & Argmax (\%) & Sample (\%) & Gain (pp) & $p$ \\
\midrule
$\mu_t^T<0.1$ & 71.00 & 71.88 & +0.88 & 0.5893 \\
$\mu_t^T<0.05$ & 65.25 & 69.00 & +3.75 & 0.0087 \\
$\mu_t^T<10^{-5}$ & 38.38 & 35.62 & -2.75 & 0.1348 \\
\bottomrule
\end{tabular*}

\end{table}
Teacher sampling improves accuracy at $\mu_t^T<0.05$, while the other two thresholds show no significant difference from argmax replacement. A problem-level Wilcoxon check supports the same conclusion at thresholds 0.1 and 0.05 ($p=0.61997$ and $p=0.01027$). These comparisons evaluate complete online policies because sampled replacements can change subsequent prefixes.

\appendixsection{After OPD}\label{app:afteropdresults}
\subsection{Accuracy and Fork Resolution across Checkpoints}\label{app:checkpoints}
\subsubsection{Original Qwen Pair}
\Cref{tab:currentcheckpointforks} reports accuracy on each checkpoint's own rollouts and fork changes on the fixed pre-training prefixes. Experimental settings are given in \cref{app:forkresolution_settings}.

\begin{table*}[!tbp]
\centering\small\setlength{\tabcolsep}{5pt}
\caption{Accuracy on each checkpoint's own rollouts and fork changes on fixed pre-training prefixes. Fork percentages use the original 729{,}574 forks as the denominator. New forks are additional positions outside the original set.}
\label{tab:currentcheckpointforks}
\begin{tabular*}{\linewidth}{@{\extracolsep{\fill}}lrrrrr@{}}
\toprule
Step & Accuracy (\%) & Solved (\%) & Unchanged (\%) & Moved (\%) & New (\%) \\
\midrule
0 & 27.10 & 0.00 & 100.00 & 0.00 & 0.00 \\
2 & 28.02 & 1.46 & 97.34 & 1.20 & 1.10 \\
4 & 28.38 & 2.38 & 95.85 & 1.76 & 1.22 \\
8 & 32.05 & 5.89 & 90.02 & 4.10 & 2.32 \\
16 & 44.12 & 14.86 & 75.18 & 9.96 & 6.32 \\
32 & 46.73 & 23.15 & 63.42 & 13.43 & 9.70 \\
48 & 46.77 & 27.36 & 57.95 & 14.69 & 11.30 \\
64 & 47.10 & 29.73 & 55.01 & 15.25 & 12.09 \\
128 & 47.52 & 32.57 & 51.87 & 15.55 & 12.29 \\
192 & 48.77 & 34.51 & 49.57 & 15.92 & 12.98 \\
256 & 48.83 & 35.48 & 48.60 & 15.92 & 12.91 \\
\bottomrule
\end{tabular*}

\end{table*}

\subsubsection{Additional Model Pairs and IFEval}\label{app:checkpointpairs}
\begin{figure*}[!tbp]
\centering
\includegraphics[width=\linewidth]{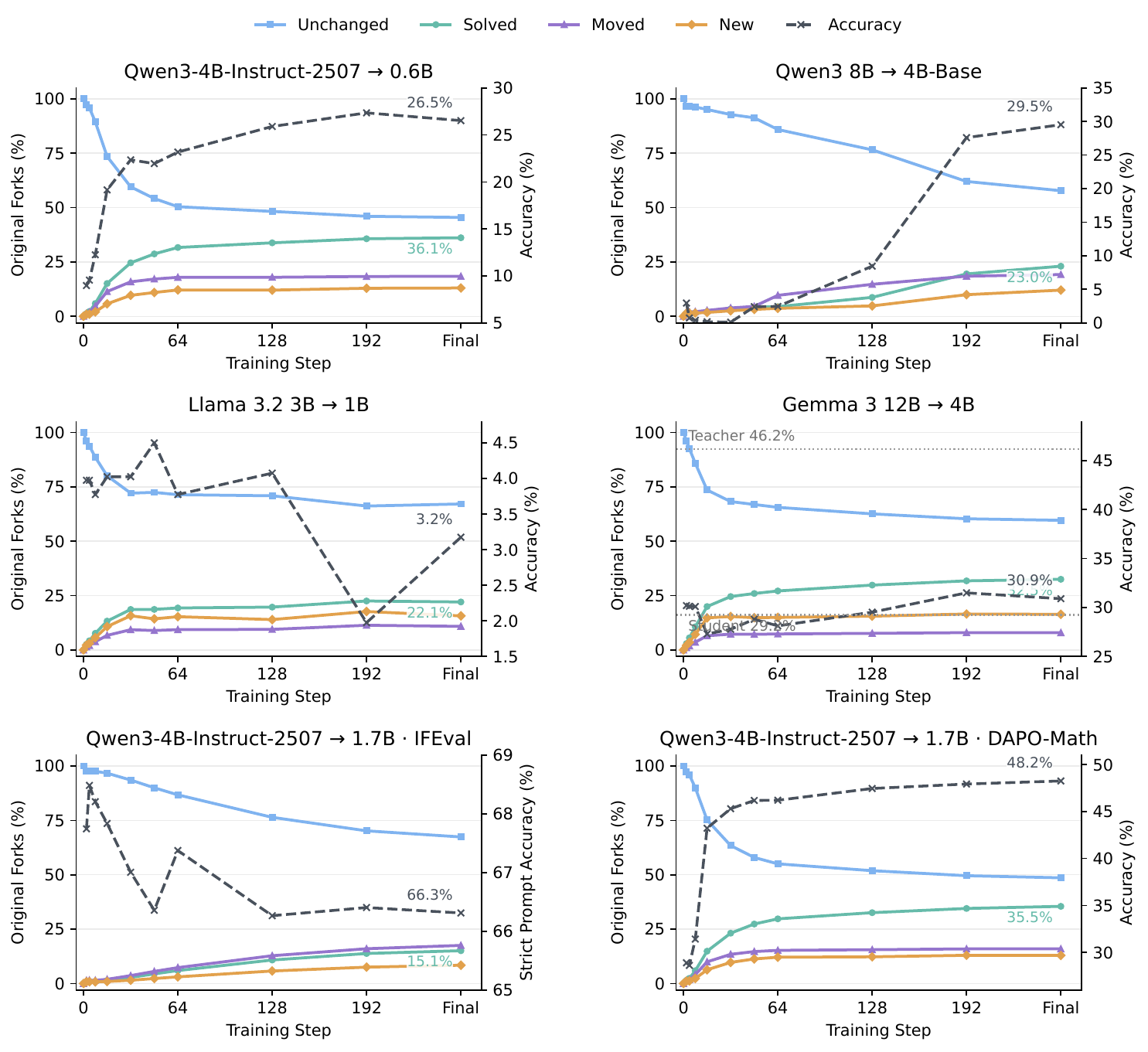}
\caption{Fork changes on fixed pre-training prefixes (left axes) and accuracy on checkpoint-generated rollouts (right axes), at temperature 0.7. Fork percentages use each pair's original fork count.}
\label{fig:afteropdadditional}
\end{figure*}

The relationship between fork resolution and accuracy differs across model pairs (\cref{fig:afteropdadditional}). In the 4B-to-0.6B pair, both improve rapidly early in training and then change more slowly. The 8B-to-4B pair has a different trajectory, with most of its accuracy increase occurring later. Thus, the timing of the gains in the original 1.7B student does not describe every OPD run.

Llama, Gemma, and IFEval show that increasing token agreement need not improve answers. By the final checkpoint, 22.1\% of original Llama forks, 32.5\% of Gemma forks, and 15.1\% of IFEval forks are solved. Their accuracy does not follow the same increase. Gemma resolves almost a third of its original forks while its final accuracy remains close to its early-checkpoint accuracy. Fork resolution therefore measures an aspect of learning that accuracy alone does not reveal, but agreement on reference prefixes is not sufficient to ensure better answers on the student's own rollouts. Evaluation settings are given in \cref{app:additionalcheckpoint_settings}.

\subsubsection{Composition of Resolved Forks}\label{app:forklearnability}
Group definitions and aggregation for \cref{fig:forklearnabilitycurrent} are given in \cref{app:learnability_settings}.

\begin{figure*}[!tbp]
\centering
\includegraphics[width=0.95\linewidth]{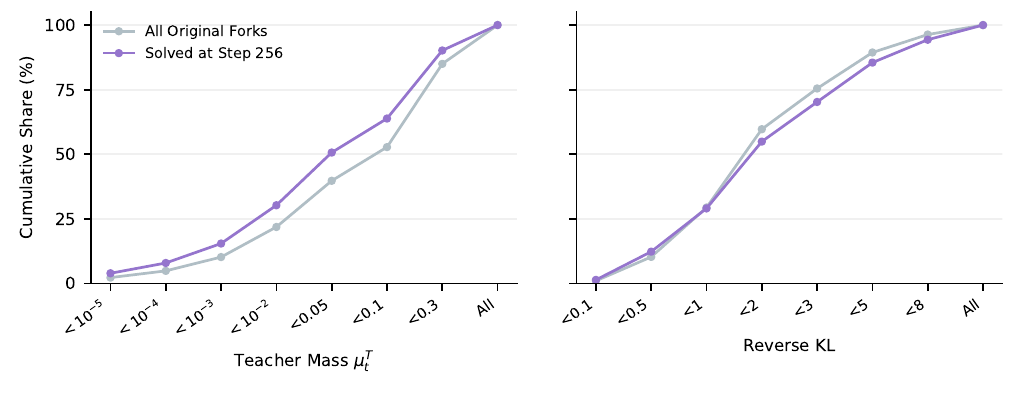}
\caption{Cumulative shares below each pre-training mass or reverse-KL threshold. The gray curve uses all 729{,}574 original forks as its denominator, and the purple curve uses the 258{,}841 forks resolved at step 256.}
\label{fig:forkresolvedcomposition}
\end{figure*}

\Cref{fig:forkresolvedcomposition} instead describes the composition of the final resolved set. At step 256, 258{,}841 original forks are resolved. Forks with $\mu_t^T<0.05$ comprise 39.7\% of the original set but 50.7\% of the resolved set. The most extreme groups remain small despite their higher resolution rates. Forks with $\mu_t^T<10^{-5}$ comprise 2.3\% of original forks and 4.0\% of resolved forks, while forks with reverse KL below 0.1 comprise 1.0\% and 1.4\%, respectively.

\subsubsection{Fork Resolution by Mass and KL across Pairs}\label{app:learnabilitypairs}

\begin{figure*}[!tbp]
\centering
\includegraphics[width=\linewidth]{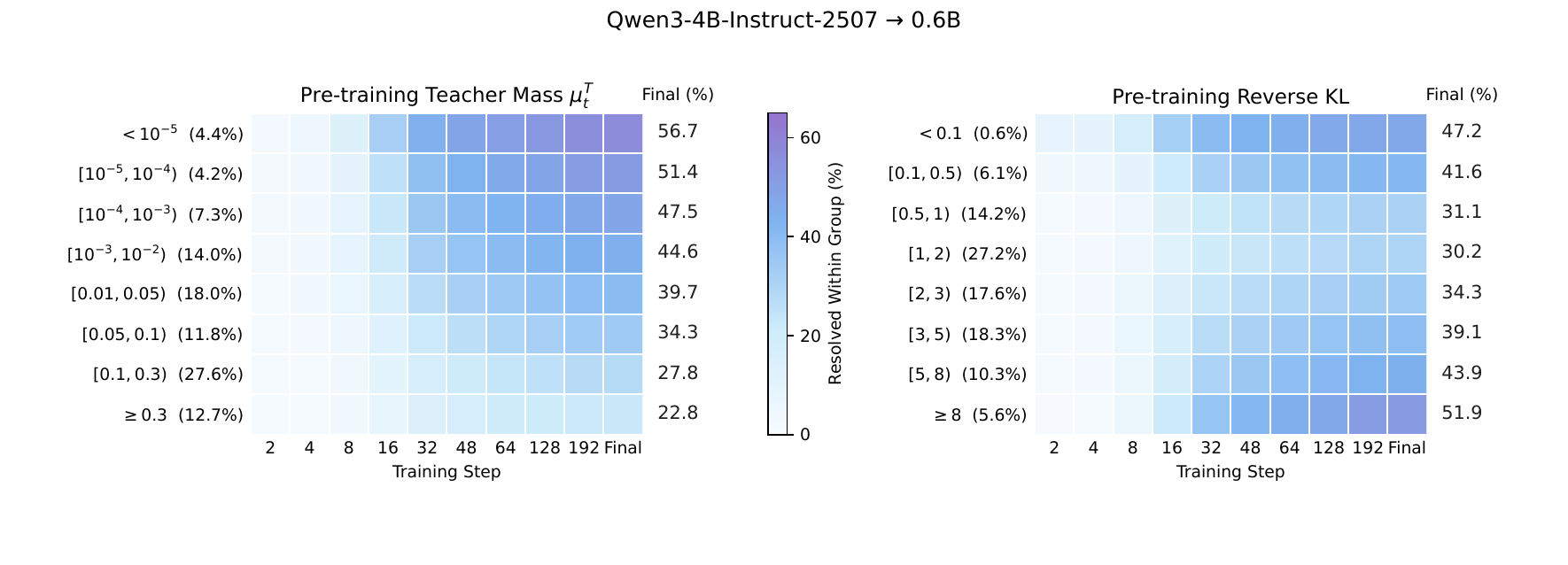}
\includegraphics[width=\linewidth]{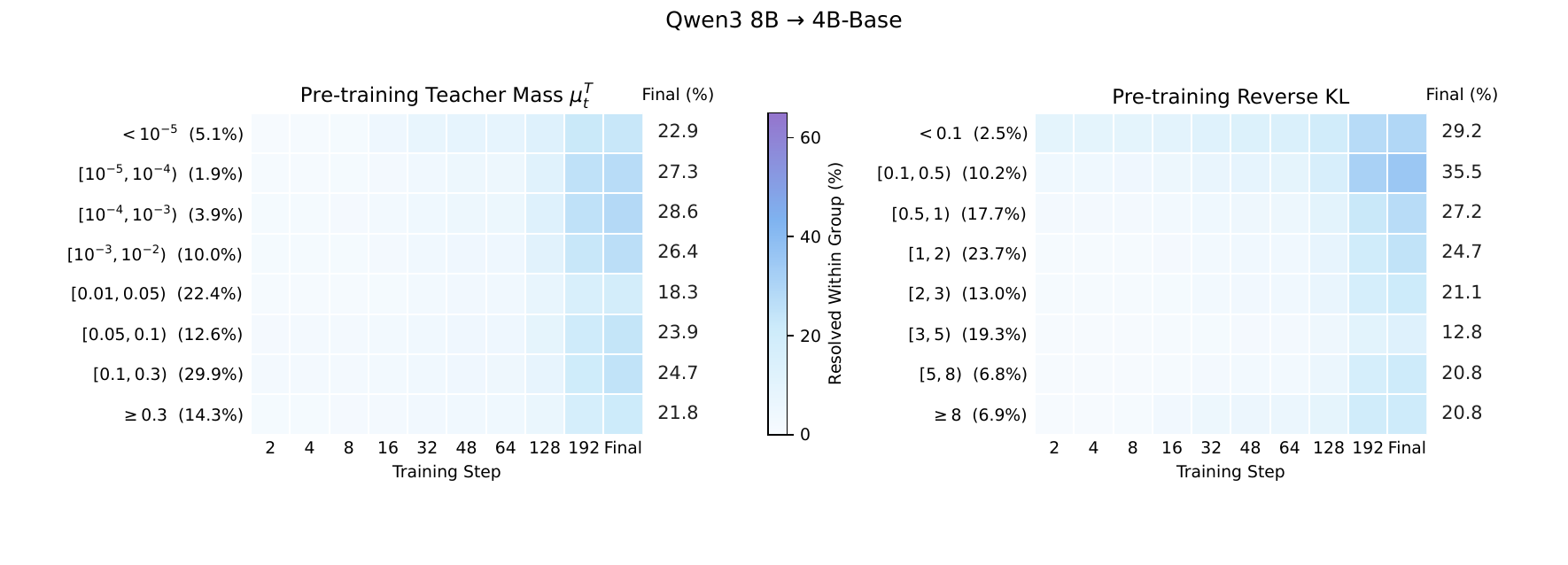}
\caption{Resolution by pre-training mass and reverse KL for the additional Qwen pairs. Colors show within-group rates, right-hand values show final rates, and parentheses show original fork shares.}
\label{fig:afteropdqwenscores}
\end{figure*}

The strong mass ordering in the original 1.7B student also appears in the 0.6B student (\cref{fig:afteropdqwenscores}). Its final resolution rate declines from 56.7\% at $\mu_t^T<10^{-5}$ to 22.8\% at $\mu_t^T\geq0.3$. The 8B-to-4B pair does not show the same monotonic ordering. Its lowest-mass group is not the most frequently resolved, and the reverse-KL profile also differs from the smaller students.

\begin{figure*}[!tbp]
\centering
\includegraphics[width=\linewidth]{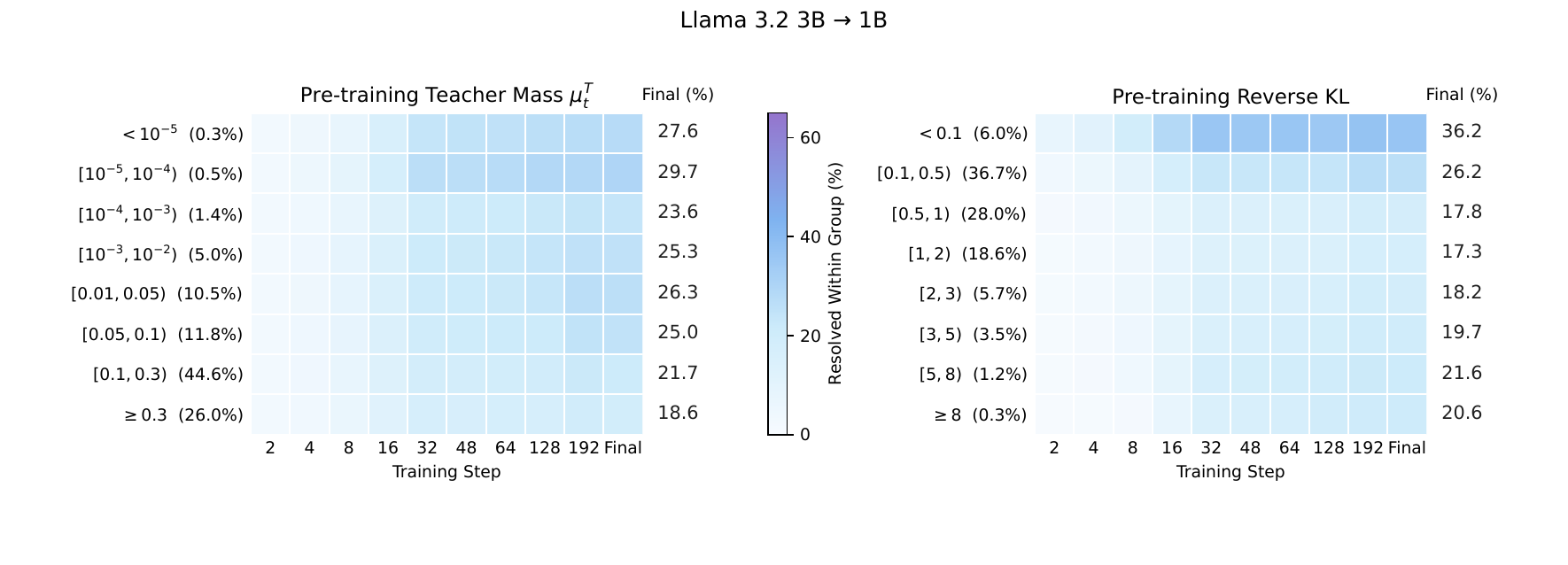}
\caption{Resolution by pre-training mass and reverse KL for Llama, using the same display and color scale as \cref{fig:afteropdqwenscores}.}
\label{fig:afteropdotherscores}
\end{figure*}

\begin{figure*}[!tbp]
\centering
\includegraphics[width=\linewidth]{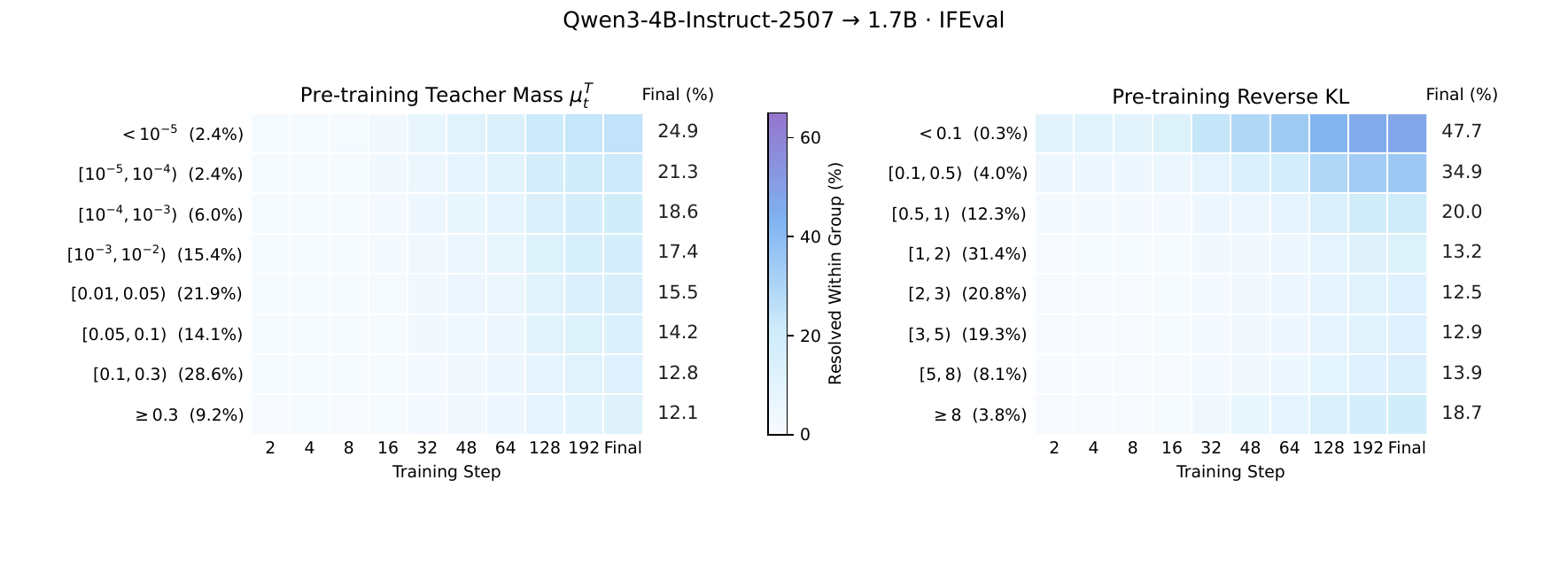}
\caption{Resolution by pre-training mass and reverse KL on IFEval, using the same display and color scale as \cref{fig:afteropdqwenscores}.}
\label{fig:afteropdifevalscores}
\end{figure*}

Lower-mass groups are more frequently resolved in Llama and IFEval, but the differences are smaller than in the 0.6B student (\cref{fig:afteropdotherscores,fig:afteropdifevalscores}). IFEval retains this ordering despite the absence of an overall accuracy increase. These profiles show why the usefulness of a selection score for online correction and its relationship to learning must be examined separately.

\subsection{KL Interventions at the Final Checkpoint}\label{app:afteropdfinalkl}
\subsubsection{Additional Qwen Model Size}
\begin{figure*}[!tbp]
\centering
\includegraphics[width=\linewidth]{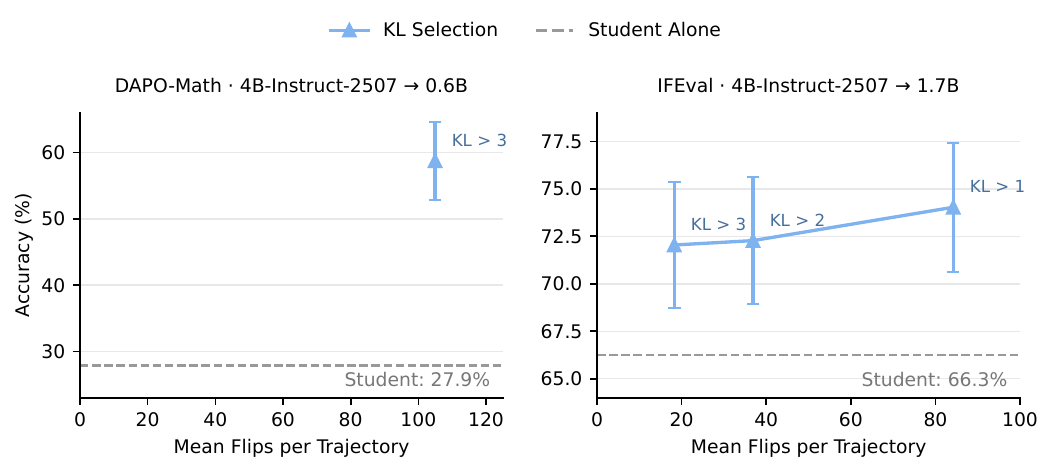}
\caption{KL-based intervention at Final on DAPO-Math (800 trajectories) and IFEval (2,164 trajectories). Dashed lines show the corresponding student-alone accuracy. Error bars are 95\% confidence intervals over problem means.}
\label{fig:afteropdfinalkl}
\end{figure*}

Teacher correction remains useful after OPD in the 4B-to-0.6B pair (\cref{fig:afteropdfinalkl}). At Final, KL-based intervention with $\mathrm{KL}>3$ raises accuracy from 27.88\% to 58.75\% with about 105 flips per trajectory. Thus, training leaves substantial accuracy recoverable through further teacher correction in this smaller student as well.

\subsubsection{Instruction Following}
Further teacher correction also improves instruction following after OPD. On IFEval, Final-checkpoint accuracy rises from 66.27\% without intervention to 72.04\% with $\mathrm{KL}>3$ and about 18 flips per trajectory. Increasing coverage with $\mathrm{KL}>1$ reaches 74.03\% with about 84 flips. These results extend the residual recovery observed in mathematics to instruction following.

\subsection{Interventions by Fork Group}\label{app:afteropdforkgroups}
\subsubsection{Original Qwen Pair}
\Cref{tab:afteropdforkgroupsfull} reports the complete current-run results. Online group assignment, calibration, and random controls are described in \cref{app:afteropdgroups_settings}.

\begin{table*}[!tbp]
\centering\footnotesize\setlength{\tabcolsep}{3pt}
\caption{Qwen fork-group interventions. F/T is flips per trajectory and F/F is the percentage of forks flipped. At step 16, New and Moved use Random (Low) and Random (Mid) as controls. Other comparisons use Random (\cref{app:stats_dualtrack}).}
\label{tab:afteropdforkgroupsfull}
\begin{tabular*}{\linewidth}{@{\extracolsep{\fill}}lrrrrrrrrrrrr@{}}
\toprule
Fork Group & \multicolumn{4}{c}{Step 16} & \multicolumn{4}{c}{Step 64} & \multicolumn{4}{c}{Final} \\
\cmidrule(lr){2-5}\cmidrule(lr){6-9}\cmidrule(lr){10-13}
 & F/T & F/F (\%) & Acc. (\%) & $p$ & F/T & F/F (\%) & Acc. (\%) & $p$ & F/T & F/F (\%) & Acc. (\%) & $p$ \\
\midrule
Random & 96.6 & 20.4 & 57.13\makebox[1.2em][l]{} & -- & 99.0 & 17.7 & 55.63\makebox[1.2em][l]{} & -- & 100.4 & 19.3 & 58.25\makebox[1.2em][l]{} & -- \\
Old & 100.0 & 19.7 & 56.75\makebox[1.2em][l]{} & 0.837 & 98.1 & 17.2 & 57.13\makebox[1.2em][l]{} & 0.353 & 99.9 & 18.3 & 56.88\makebox[1.2em][l]{} & 0.396 \\
Unchanged & 94.5 & 18.5 & \textbf{57.63}\makebox[1.2em][l]{} & 0.761 & 97.6 & 17.0 & \textbf{58.38}\makebox[1.2em][l]{} & 0.086 & 100.4 & 18.6 & \textbf{58.50}\makebox[1.2em][l]{} & 0.878 \\
\midrule
Random (Low) & 28.9 & 5.9 & 51.00\makebox[1.2em][l]{} & -- & -- & -- & -- & -- & -- & -- & -- & -- \\
New & 31.4 & 6.8 & 44.88\makebox[1.2em][l]{$^{***}$} & $<0.001$ & 78.0 & 13.8 & 49.50\makebox[1.2em][l]{$^{***}$} & $<0.001$ & 82.5 & 15.9 & 54.63\makebox[1.2em][l]{$^{*}$} & 0.024 \\
\midrule
Random (Mid) & 52.2 & 10.8 & 50.88\makebox[1.2em][l]{} & -- & -- & -- & -- & -- & -- & -- & -- & -- \\
Moved & 45.7 & 9.0 & 51.13\makebox[1.2em][l]{} & 0.885 & 86.5 & 15.7 & 57.38\makebox[1.2em][l]{} & 0.292 & 90.9 & 17.4 & 57.13\makebox[1.2em][l]{} & 0.488 \\
\bottomrule
\end{tabular*}

\end{table*}

The $p$ values in \cref{tab:afteropdforkgroupsfull} use two-sided paired $t$-tests over the 200 problem means, each averaging four outcomes (\cref{app:stats_dualtrack}). These exploratory comparisons are unadjusted.

\subsubsection{Gemma and Llama}\label{app:afteropdfamilygroups}
New and Moved selection provide limited recovery in Gemma across the tested checkpoints (\cref{tab:afteropdgroupsgemma}). At final, their accuracies are 32.5\% and 30.5\%, compared with 32.0\% without intervention. Old and Unchanged selection instead reach 43.0\%, using more flips. Under these policies, most of the observed recovery comes from correcting forks that retain the base student's preference rather than newly introduced or moved forks.

New and Moved selection also remain close to student-alone accuracy in Llama (\cref{tab:afteropdgroupsllama}). At final, both reach 3.5\%, compared with 1.6\% without intervention and 6.0\% under Unchanged selection. The gains are smaller than in Gemma, and the group ranking varies across checkpoints.

\begin{table*}[!tbp]
\centering\footnotesize\setlength{\tabcolsep}{3pt}
\caption{Gemma 12B-to-4B fork-group interventions on 200 problems. F/T and F/F follow \cref{tab:afteropdforkgroupsfull}. $p$ compares each policy's one outcome per problem with the mean of four Student Alone outcomes using a two-sided paired $t$-test.}
\label{tab:afteropdgroupsgemma}
\begin{tabular*}{\linewidth}{@{\extracolsep{\fill}}lrrrrrrrrrrrr@{}}
\toprule
Fork Group & \multicolumn{4}{c}{Step 16} & \multicolumn{4}{c}{Step 64} & \multicolumn{4}{c}{Final} \\
\cmidrule(lr){2-5}\cmidrule(lr){6-9}\cmidrule(lr){10-13}
 & F/T & F/F (\%) & Acc. (\%) & $p$ & F/T & F/F (\%) & Acc. (\%) & $p$ & F/T & F/F (\%) & Acc. (\%) & $p$ \\
\midrule
Student Alone & 0.0 & 0.0 & 26.75\makebox[1.2em][l]{} & -- & 0.0 & 0.0 & 27.63\makebox[1.2em][l]{} & -- & 0.0 & 0.0 & 32.00\makebox[1.2em][l]{} & -- \\
\midrule
New & 15.2 & 14.6 & 29.00\makebox[1.2em][l]{} & 0.297 & 16.6 & 16.7 & 29.00\makebox[1.2em][l]{} & 0.531 & 17.0 & 18.8 & 32.50\makebox[1.2em][l]{} & 0.849 \\
Moved & 7.2 & 6.7 & 29.00\makebox[1.2em][l]{} & 0.333 & 8.4 & 8.3 & 28.50\makebox[1.2em][l]{} & 0.711 & 8.8 & 9.1 & 30.50\makebox[1.2em][l]{} & 0.509 \\
Unchanged & 74.0 & 74.3 & \textbf{44.50}\makebox[1.2em][l]{$^{***}$} & $<0.001$ & 66.5 & 70.4 & \textbf{39.50}\makebox[1.2em][l]{$^{***}$} & $<0.001$ & 60.5 & 67.8 & \textbf{43.00}\makebox[1.2em][l]{$^{***}$} & $<0.001$ \\
Old & 78.7 & 80.1 & 41.00\makebox[1.2em][l]{$^{***}$} & $<0.001$ & 72.7 & 77.6 & 39.00\makebox[1.2em][l]{$^{***}$} & $<0.001$ & 66.8 & 77.0 & \textbf{43.00}\makebox[1.2em][l]{$^{***}$} & $<0.001$ \\
\bottomrule
\end{tabular*}

\end{table*}

\begin{table*}[!tbp]
\centering\footnotesize\setlength{\tabcolsep}{3pt}
\caption{Llama 3B-to-1B fork-group interventions, using the same columns and problem-paired comparison with Student Alone as \cref{tab:afteropdgroupsgemma}.}
\label{tab:afteropdgroupsllama}
\begin{tabular*}{\linewidth}{@{\extracolsep{\fill}}lrrrrrrrrrrrr@{}}
\toprule
Fork Group & \multicolumn{4}{c}{Step 16} & \multicolumn{4}{c}{Step 64} & \multicolumn{4}{c}{Final} \\
\cmidrule(lr){2-5}\cmidrule(lr){6-9}\cmidrule(lr){10-13}
 & F/T & F/F (\%) & Acc. (\%) & $p$ & F/T & F/F (\%) & Acc. (\%) & $p$ & F/T & F/F (\%) & Acc. (\%) & $p$ \\
\midrule
Student Alone & 0.0 & 0.0 & 3.63\makebox[1.2em][l]{} & -- & 0.0 & 0.0 & 2.63\makebox[1.2em][l]{} & -- & 0.0 & 0.0 & 1.63\makebox[1.2em][l]{} & -- \\
\midrule
New & 10.4 & 12.1 & 3.50\makebox[1.2em][l]{} & 0.921 & 36.9 & 24.8 & 4.00\makebox[1.2em][l]{} & 0.289 & 23.6 & 16.1 & 3.50\makebox[1.2em][l]{} & 0.116 \\
Moved & 5.3 & 6.2 & 3.50\makebox[1.2em][l]{} & 0.927 & 23.2 & 9.3 & 4.00\makebox[1.2em][l]{} & 0.228 & 26.7 & 9.4 & 3.50\makebox[1.2em][l]{} & 0.116 \\
Unchanged & 77.1 & 55.6 & 3.00\makebox[1.2em][l]{} & 0.620 & 107.0 & 39.5 & 4.00\makebox[1.2em][l]{} & 0.289 & 94.1 & 43.7 & \textbf{6.00}\makebox[1.2em][l]{$^{**}$} & 0.005 \\
Old & 84.1 & 58.7 & \textbf{6.00}\makebox[1.2em][l]{} & 0.149 & 98.3 & 41.7 & \textbf{5.00}\makebox[1.2em][l]{} & 0.117 & 87.5 & 66.0 & 4.50\makebox[1.2em][l]{} & 0.051 \\
\bottomrule
\end{tabular*}

\end{table*}

Gemma's remaining Old forks still admit useful correction at later checkpoints. Selecting Old forks reaches 41.0\% accuracy at step 16 and 43.0\% at final, with about 79 and 67 flips per trajectory (\cref{tab:afteropdgroupsgemma}). The persistence of this recovery contrasts with the limited accuracy improvement during training. Llama does not show a comparably strong recovery under these group policies (\cref{tab:afteropdgroupsllama}). The benefit of post-training intervention must therefore be measured for each pair rather than inferred from the fraction of forks resolved.

\subsubsection{IFEval}\label{app:ifevalafteropdgroups}
None of the four group policies yields a consistent accuracy improvement across IFEval checkpoints at the tested low flip counts (\cref{tab:afteropdgroupsifeval}). Differences from student-alone accuracy range from $-1.06$ to $+0.65$ pp, with 0.44 to 0.58 flips per trajectory. New and Moved selection therefore add little recovery in this setting, but the same is true of Old and Unchanged selection at these counts.

\begin{table*}[!tbp]
\centering\footnotesize\setlength{\tabcolsep}{3pt}
\caption{IFEval fork-group interventions, with 2{,}164 trajectories per condition. F/T is flips per trajectory. Accuracy uses strict prompt scoring.}
\label{tab:afteropdgroupsifeval}
\begin{tabular*}{\linewidth}{@{\extracolsep{\fill}}lrrrrrr@{}}
\toprule
Rule & \multicolumn{2}{c}{Step 16} & \multicolumn{2}{c}{Step 64} & \multicolumn{2}{c}{Final} \\
\cmidrule(lr){2-3}\cmidrule(lr){4-5}\cmidrule(lr){6-7}
 & F/T & Acc. (\%) & F/T & Acc. (\%) & F/T & Acc. (\%) \\
\midrule
Student Alone & 0.00 & 67.38 & 0.00 & 68.99 & 0.00 & 65.48 \\
\midrule
New & 0.48 & 67.51 & 0.47 & 68.67 & 0.50 & \textbf{66.13} \\
Moved & 0.44 & 67.47 & 0.49 & 67.93 & 0.50 & 65.57 \\
Unchanged & 0.55 & \textbf{67.98} & 0.58 & \textbf{69.13} & 0.56 & 65.53 \\
Old & 0.56 & 67.51 & 0.56 & 68.21 & 0.51 & 65.53 \\
\bottomrule
\end{tabular*}

\end{table*}

\clearpage
\subsection{Distribution Changes Beyond Fork Resolution}\label{app:afteropdentropy}
\subsubsection{Why Might More Resolved Forks Fail to Improve Accuracy?}
Fork resolution increases in Llama, Gemma, and IFEval without a corresponding increase in accuracy (\cref{fig:afteropdadditional}). Resolving an argmax disagreement changes the student's preferred token, while generation depends on the full token distribution. The usefulness of the acquired preferences and accompanying changes in token probabilities are therefore separate questions when agreement increases without better answers.

\begin{figure*}[!tbp]
\centering
\includegraphics[width=\linewidth,height=0.72\textheight,keepaspectratio]{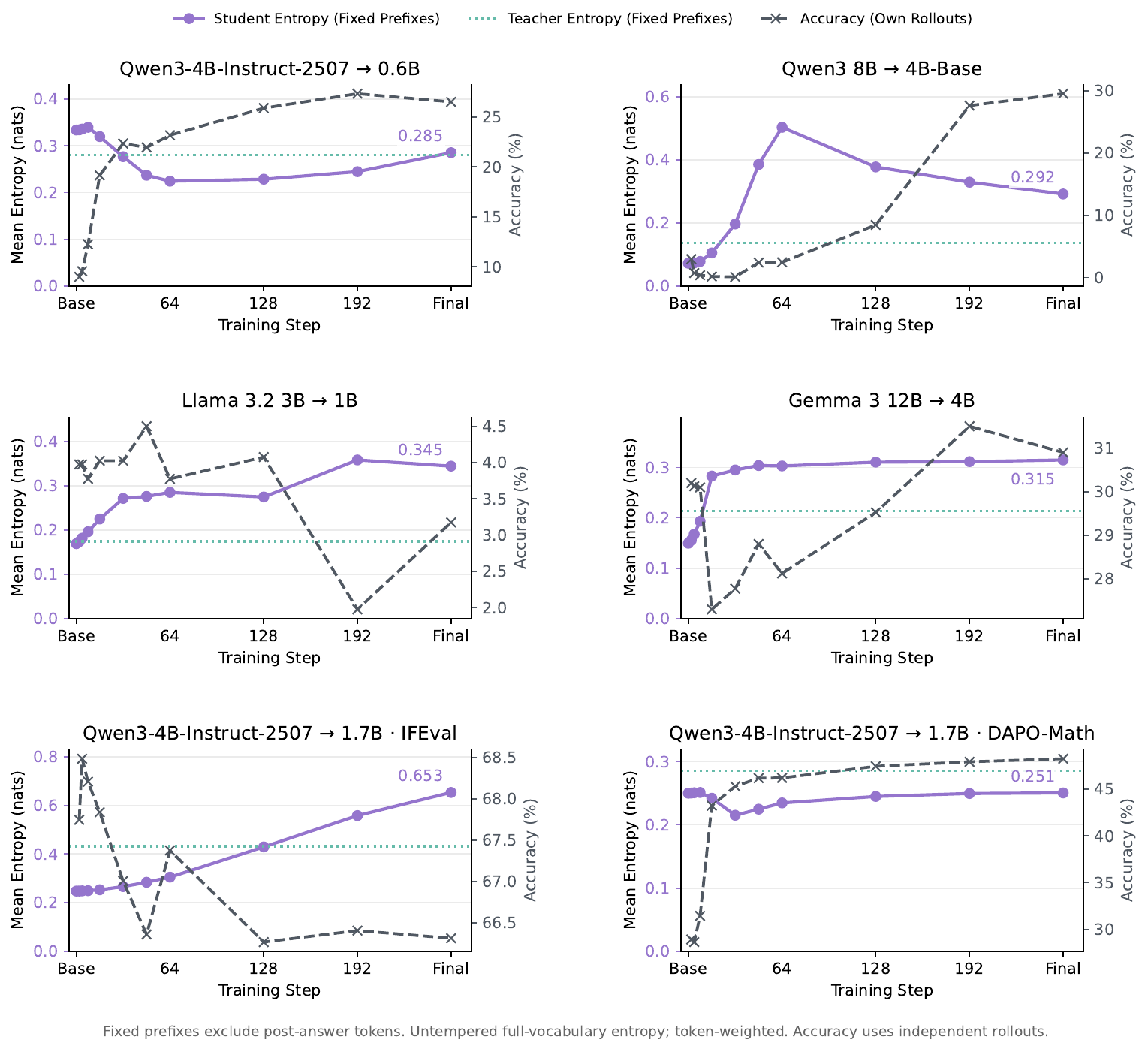}
\caption{Student entropy on fixed pre-training prefixes and accuracy on checkpoint-generated rollouts. Post-answer repetition is excluded from entropy scoring. Settings are in \cref{app:additionalcheckpoint_settings}.}
\label{fig:afteropdentropy}
\end{figure*}

Mean entropy on the same pre-training prefixes rises substantially in these three settings (\cref{fig:afteropdentropy}). From Base to Final, it rises from 0.169 to 0.345 nats in Llama, from 0.150 to 0.315 in Gemma, and from 0.248 to 0.653 on IFEval. The original 1.7B Qwen student's entropy remains near 0.250 nats, while the 0.6B student's changes from 0.334 to 0.285. Both Qwen students improve in accuracy. The 8B-to-4B pair follows a different trajectory, with final entropy above Base but late accuracy gains accompanying a decline from its intermediate entropy peak.

These entropy trajectories motivate examining whether training changes the distribution at positions where the student already agrees with the teacher. Such changes are invisible to fork resolution because the preferred token can remain the same while its probability changes.

\subsubsection{Probability Changes at Originally Agreed Positions}
We retain the same pre-answer positions where the base student and teacher share an argmax. Original Non-Forks includes all these positions. Still Agree further requires the final checkpoint to retain that shared argmax. Each row compares Base and Final on exactly the same positions. Entropy uses the full vocabulary, and probabilities are untempered. When the teacher's token falls outside the saved top-16, its probability is bounded by the sixteenth token's probability rather than set to zero.

\begin{table*}[!tbp]
\centering\footnotesize\setlength{\tabcolsep}{3pt}
\caption{Paired distribution changes from Base to Final at originally agreed positions. $H$ is entropy in nats, $P$ is the probability of the teacher's argmax in percent, and $N$ is the number of positions in millions. Still Agree uses positions where all three argmax choices agree.}
\label{tab:nonforkdistribution}
\begin{tabular*}{\linewidth}{@{\extracolsep{\fill}}llrrrrrrr@{}}
\toprule
Pair / Dataset & Positions & $N$ (M) & $H_{\rm Base}$ & $H_{\rm Final}$ & $\Delta H$ & $P_{\rm Base}$ & $P_{\rm Final}$ & $\Delta P$ (pp) \\
\midrule
Qwen 4B--1.7B & Original Non-Forks & 4.45 & 0.135 & 0.124 & -0.011 & 95.49 & 95.32 & -0.17 \\
 & Still Agree & 4.36 & 0.119 & 0.105 & -0.014 & 96.13 & 96.80 & +0.68 \\
\midrule
Qwen 4B--0.6B & Original Non-Forks & 2.52 & 0.183 & 0.126 & -0.058 & 94.17 & 95.35 & +1.18 \\
 & Still Agree & 2.46 & 0.162 & 0.101 & -0.061 & 94.96 & 97.21 & +2.25 \\
\midrule
Qwen 8B--4B & Original Non-Forks & 9.39 & 0.031 & 0.108 & +0.077 & 99.17 & 97.90 & -1.27 \\
 & Still Agree & 9.36 & 0.028 & 0.100 & +0.072 & 99.28 & 98.19 & -1.08 \\
\midrule
Llama 3B--1B & Original Non-Forks & 6.87 & 0.120 & 0.264 & +0.144 & 96.70 & 94.39 & -2.32 \\
 & Still Agree & 6.83 & 0.112 & 0.251 & +0.139 & 96.99 & 94.79 & -2.20 \\
\midrule
Gemma 12B--4B & Original Non-Forks & 5.86 & 0.105 & 0.243 & +0.138 & 96.38 & 92.49 & -3.89 \\
 & Still Agree & 5.80 & 0.097 & 0.230 & +0.133 & 96.75 & 93.20 & -3.54 \\
\midrule
Qwen 1.7B / IFEval & Original Non-Forks & 0.97 & 0.098 & 0.311 & +0.213 & 96.82 & 91.99 & -4.83 \\
 & Still Agree & 0.95 & 0.082 & 0.279 & +0.197 & 97.44 & 93.24 & -4.20 \\
\bottomrule
\end{tabular*}

\end{table*}

Entropy increases even on Still Agree positions, from 0.112 to 0.251 nats in Llama, 0.097 to 0.230 in Gemma, and 0.082 to 0.279 on IFEval. The shared token's untempered probability falls by 2.20, 3.54, and 4.20 pp in these settings. These changes therefore extend beyond newly introduced forks. The original 1.7B and 0.6B Qwen students instead show lower entropy and higher shared-token probability on these positions. The 8B-to-4B pair improves accuracy despite higher final entropy and lower shared-token probability than Base. Its late accuracy gains accompany a decline from the intermediate entropy peak, making the trajectory of distributional change more informative than the Base-to-Final difference alone.

\end{document}